\documentclass{article} 
\usepackage{iclr2025_conference,times}

\usepackage{amsmath,amsfonts,bm}

\def\eqref#1{equation~\ref{#1}}

\def\1{\bm{1}}

\DeclareMathAlphabet{\mathsfit}{\encodingdefault}{\sfdefault}{m}{sl}
\SetMathAlphabet{\mathsfit}{bold}{\encodingdefault}{\sfdefault}{bx}{n}

\usepackage{hyperref}
\usepackage{url}
\usepackage{adjustbox}
\usepackage{tabularx}
\usepackage{booktabs}
\usepackage{multirow}
\usepackage{amssymb}
\usepackage[ruled,vlined]{algorithm2e}
\usepackage{algorithmic}
\usepackage{tcolorbox}
\usepackage{seqsplit}
\usepackage{xcolor}
\usepackage{inconsolata}
\usepackage{paralist, tabularx}
\usepackage{cleveref}
\usepackage{wrapfig}
\usepackage{paralist, tabularx}
\usepackage{enumitem}

\definecolor{headercolor}{RGB}{200, 200, 255}
\definecolor{contentcolor}{RGB}{245, 245, 245}

\title{Large Language Model is Secretly a Protein Sequence Optimizer}

\author{Yinkai Wang\footnotemark[1] \\
Tufts University \\
\And
Jiaxing He \\
Northeastern University\\
\And
Yuanqi Du \\
Cornell University\\
\And
Xiaohui Chen\\
Tufts University\\
\And 
Jianan Canal Li \\
UC Berkeley \\
\And 
Li-Ping Liu\\
Tufts University\\
\And
Xiaolin Xu\\
Northeastern University\\
\And
Soha Hassoun\footnotemark[1]\\
Tufts University\\
}

\iclrfinalcopy 
\begin{document}

\maketitle

\footnotetext[1]{Correspondence to: Yinkai Wang $<$yinkai.wang@tufts.edu$>$ and Soha Hassoun $<$soha@cs.tufts.edu$>$.}

\begin{abstract}
We consider the protein sequence engineering problem, which aims to find protein sequences with high fitness levels, starting from a given wild-type sequence. Directed evolution has been a dominating paradigm in this field which has an iterative process to generate variants and select via experimental feedback. We demonstrate large language models (LLMs), despite being trained on massive texts, are secretly protein sequence optimizers. With a directed evolutionary method, LLM can perform protein engineering through Pareto and experiment-budget constrained optimization, demonstrating success on both synthetic and experimental fitness landscapes. 
\end{abstract}
\vspace{-3mm}
\vspace{-0.5em}
\section{Introduction}
\vspace{-0.5em}

Protein engineering aims to develop novel protein sequences exhibiting improved or new-to-nature  functions~\citep{romero2009exploring}. \textit{Directed evolution} stands as a cornerstone paradigm of the field which leverages iterative rounds of mutagenesis and experimental selection to yield variants with gradually enhanced fitness~\citep{arnold1998design}. While classical directed evolution has proven effective, it is generally acknowledged that its greedy optimization process often converges on suboptimal variants once a local maximum in the sequence fitness landscape of activity is reached~\citep{yang2019machine}. In recently proposed machine-learning guided directed evolution (MLDE) settings, sequence-to-function models have been incorporated as a computational surrogate to select candidates for experimental validation~\citep{yang2019machine,kirjner2023improving,ren2022proximal,jain2022biological,brookes2019conditioning,yang2024active}.

With the grand success of AlphaFold2 on accurately predicting protein tertiary structures~\citep{jumper2021highly}, numerous work study protein language models (PLMs) which do not rely on multiple sequence alignment (MSA) and instead counting on learning the co-evolution information from multi-head attention transformers~\citep{lin2023evolutionary, zhang2024protein}. Motivated by the improved performance on structure prediction emerged from sequence-based pre-training, it has been employed as part of an evolutionary method which designs two masking strategies as mutation operators \citep{tran2024protein}. More recently, increasing attention has been attracted to leverage large language models (LLMs) for problems in scientific discovery, e.g. molecule optimization~\citep{wang2024efficient}, materials discovery~\citep{lu2024generative}. In protein engineering, \citet{chen2024llms} propose a bi-level optimization to iteratively fine-tune pre-trained LLMs for protein optimization.

In this paper, we demonstrate LLMs themselves can already optimize protein fitness on-the-fly without further fine-tuning. Specifically, we build an evolutionary method that directly samples from pre-trained LLMs and select high fitness and low editing distance candidates for the next iteration. We count on LLMs to propose new candidates (i.e. mutation and crossover) to guide the search. Upon multiple experiments from 1) experiment-derived exact fitness landscapes, 2) simulated synthetic fitness landscapes, and 
 3) machine learning (ML) fitness landscape models trained on deep mutational scanning (DMS) datasets, we demonstrate LLMs can effectively propose new candidates that are much more efficient than the straightforward evolutionary algorithm with random mutation and crossover. We also extend the experiment setting to experiment-budget constrained and multi-objective optimization.

\vspace{-0.5em}
\section{Preliminary: Protein Sequence Optimization}
\vspace{-0.5em}

\textbf{Single-objective optimization.} Given an oracle function $f:\Omega \rightarrow \mathbb{R}$, where $\Omega := \{(a_1, a_2, \cdots, a_L) | a_i \in \mathcal{A}\}$, $L$ is the maximum length of a protein sequence, and $\mathcal{A}$ is the set of 20 amino acid types, we aim to find the candidate $x^*$ as follows:
\begin{equation}
x^* = \arg\max_{x \in \Omega} f(x)
\end{equation}
\vspace{-1em}

\textbf{Constrained optimization.} Beyond merely optimizing the oracle function, we are often limited by experimental budget such that we constrain the maximum number of edits to be $K$ from the reference wild type $x_{\text{ref}}$.
\begin{align}
x^* = \arg\max_{x \in \Omega} f(x),~\text{s.t.} \; \text{dist}(x, x_{\text{ref}}) \leq K
\end{align}
where the distance function is taken as the Hamming distance $d_H(\cdot, \cdot)$ between two sequences.

\textbf{Budget-constrained optimization.} Instead of constraining the absolute Hamming distance, a more realistic setting in wet-lab experiments is to constrain the relative Hamming distance (i.e. minimum Hamming distance between the proposed sequence and all previous experiment trials).
\begin{equation}
\text{dist}(x, \mathcal{P}) = \min_{x_p \in \mathcal{P}} d_H(x, x_p)
\end{equation}
where $\mathcal{P}$ is the set of all previously evaluated candidates.

\textbf{Multi-objective optimization.} In scenarios where we have multiple oracle functions to optimize, we solve a multi-objective optimization problem where the objective function becomes a vector-valued function $f: \Omega \rightarrow \mathbb{R}^d$:
\begin{equation}
x^* = \arg\max_{x \in \Omega} f(x)
\end{equation}
One simple way to aggregate multiple objectives is to take a weighted sum over the output vector $\sum_j w_j f_j(x)$ and $\sum_j w_j = 1$, where we refer to as \textit{sum of objectives}.

Nevertheless, the more rigorous formulation is to find the Pareto frontier $\mathcal{P}$, defined as follows:
\begin{equation}
\mathcal{P}(\mathcal{X}) = \big\{ x \in \mathcal{X} :\{ x' \in \mathcal{X}: x \preceq x', x \neq x' \} = \varnothing   \big\}
\end{equation}
where $\preceq$ defines a partial order such that $x \preceq x'$ or $x$ is dominated by $x'$ if and only if $\forall_j\; f_j(x') \geq f_j(x)$. We refer the problem to find the Pareto set to as \textit{Pareto set selection}.

\begin{figure}[t]
    \centering
    \includegraphics[width=\linewidth]{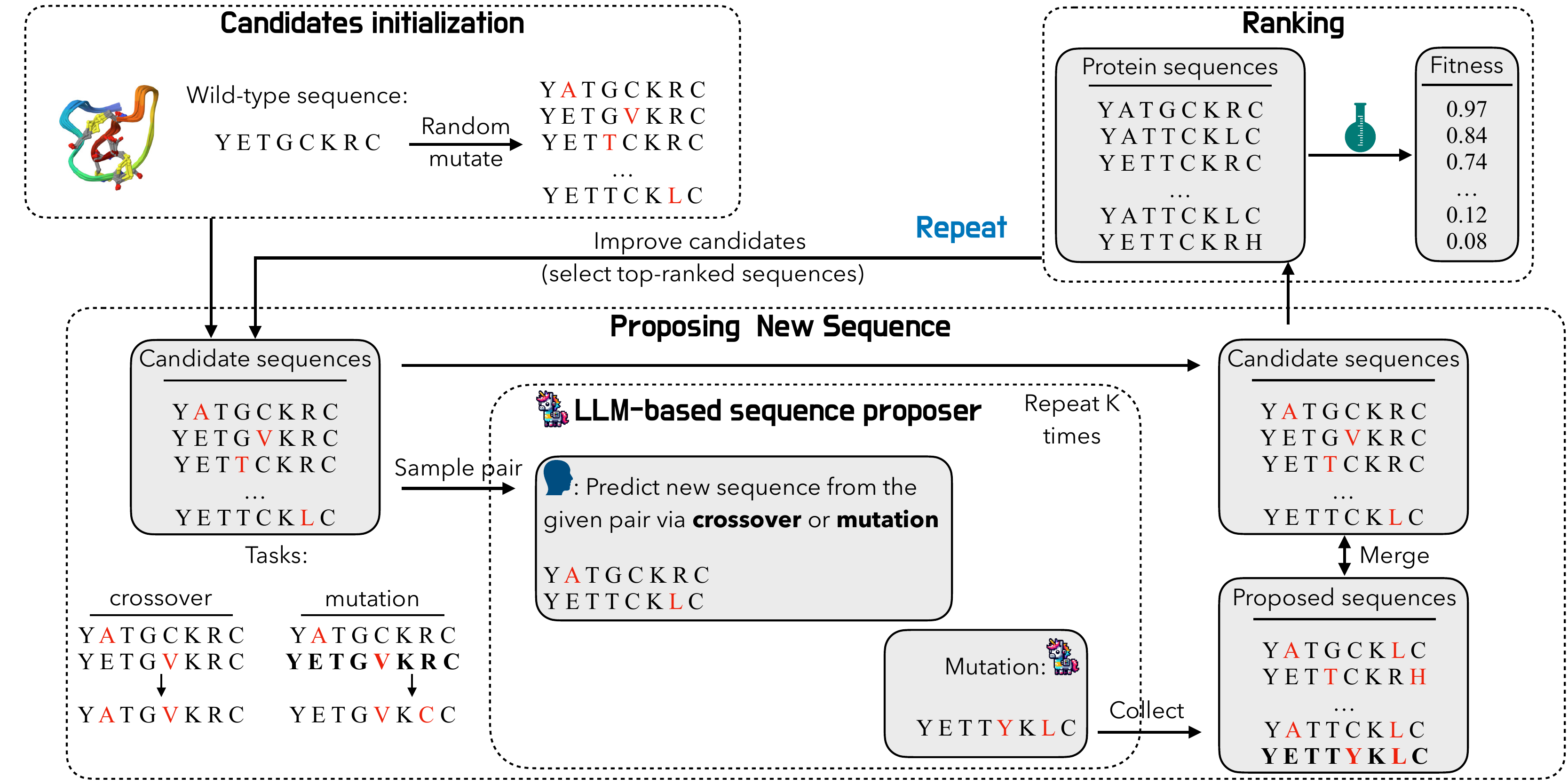}
\vspace{-2em}
    \caption{The overview of the optimization framework. }
    \label{fig:main}
\vspace{-1em}
\end{figure}

\vspace{-0.5em}
\section{Methodology}\label{sec:method}
\vspace{-0.5em}

We propose an evolutionary method for protein sequence optimization. There are three main modules in our method: (1) initialization, (2) diversification and (3) selection. The framework is illustrated in \Cref{fig:main} and the pseudocode is included in \Cref{alg:protein_opt}.

\textbf{Initialization.} We initialize a pool of candidates by randomly sampling from the entire space or a single mutation from the wild type. 

\textbf{Mutation/Crossover.} The default mutation in evolutionary algorithm (EA) is to perform a random mutation over a single protein sequence; the default crossover in EA is to randomly swap amino acids of two protein sequences at the same position or swap the entire half sequence split by a random position. In our LLM-based method, we randomly sample a pair of protein sequences from our pool and encourage LLMs to propose a new candidate either through mutation or crossover.

\textbf{Selection.} For single-objective optimization, we simply select the top-k ranked protein sequences in both the previous pool and the newly proposed candidates. For constrained optimization, we employ a rejection sampling-based strategy: we discard all samples that violate the constraints (exceeding the maximum number of edits allowed. For multi-objective optimization, we optimize for two objectives: (1) \textbf{objective scalarization}: we sum over all objective values in the multi-objective vectors and treat it as a single-objective optimization problem; (2) \textbf{Pareto set selection}: we select only the candidates on the Pareto frontier to proceed the next iteration.

\vspace{-0.5em}
\section{Experiment}
\vspace{-0.5em}

\begin{table*}[t]
\centering
\resizebox{\textwidth}{!}{
\begin{tabular}{lccccccc}\toprule
 Dataset & Space size  & Sequence length & \# mutation sites & Oracle & Target range & Initial pool fitness & wild-type fitness \\
\toprule
Syn-3bfo & N/A & 85& 85& SLIP & N/A &-4.11$\pm$2.22& 0.00\\ 
\midrule 
GB1 & 149,361& 56 &4& exact & $[0,8.76]$ & 0.08$\pm$0.40&1.00 \\
TrpB & 159,129& 397 &4& exact & $[0,1]$ & 0.02$\pm$0.06&0.41 \\
\midrule 
AAV & N/A &735 & 28 & ML & N/A & 0.31$\pm$0.05 & 0.56\\
GFP & N/A &238 & 237 & ML & N/A & 0.08$\pm$0.12 &0.94\\ 
\bottomrule
 \end{tabular}}

\vspace{-0.5em}
\caption{Dataset statistics. Syn-3bfo refer to synthetic dataset constructed from PDB ID 3bfo.} 
\vspace{-0.5em}

\label{tab:data_stat}
\vspace{-1em}
\end{table*}

\vspace{-0.5em}
\subsection{Experiment Set-up}
\vspace{-0.5em}

\textbf{Oracle function.} We have three types of oracle functions: \textbf{exact oracle}, \textbf{synthetic SLIP model oracle}, and \textbf{ML oracle}. For \textbf{exact oracle}, directly measures the fitness values of all possible variants in a specified search space by deep mutational scanning (DMS)~\citep{fowler2014deep}. Due to experimental budget constraints, the number of sites to be mutated is often limited to four or fewer. 

For the \textbf{synthetic SLIP oracle}, the statistical energy of protein variants evaluated by the Potts model has been demonstrated to correlate with observed empirical fitnes~\citep{hopf2017mutation}, and the Synthetic Landscape Inference for Proteins (SLIP) based on Potts models has been proposed as a hard-to-optimize fitness landscape~\citep{thomas2022tuned}.  

For \textbf{ML oracle}, a machine learning model is trained on sequence–fitness pairs of single and multiple mutants for a wild-type protein through DMS ~\citep{dallago2021flip}. Unlike the exact oracle, which focuses on a small subset of variants, the ML oracle can evaluate protein variants with any number of mutations away from the wild-type sequence, returning a predicted fitness value. However, its generalization ability remains a key limitation: because the model is typically trained on a comparatively small dataset, its predictions may be unreliable across the full sequence space.

\textbf{Hyperparameters.} We adopt the \textit{Llama-3.1-8B-Instruct} model as our LLM. To mimic real-world protein engineering experiment procedure, we choose a set of 32/48/96 candidates in each iteration for a total of 4 iterations. For experiment settings allowing a larger number of mutations away from the wild-type, we increase to 8 iterations for better optimization.

\textbf{Baselines.} We use the exactly same hyperparameters and initial pools for the baseline evolutionary algorithm as our model, we adopt the default mutation and crossover operators for EA in~\Cref{sec:method}.

\textbf{Datasets.} Here we list the datasets used for each type of oracle function:
\begin{itemize}[leftmargin=1.5em]

\vspace{-0.5em}
    \item For the exact oracle setting, we test our framework on two combinatorial landscape datasets GB1~\citep{wu2016adaptation} and TrpB~\citep{johnston2024combinatorially}. On these landscapes, four amino acids are picked to be mutated, therefore having a total of $20^{4}$ variants. The fitness is measured by wet-lab experiments for nearly all the variants in the library. 
    \item For the synthetic SLIP oracle setting, we create the tuned synthetic landscape constructed from the multiple sequence alignment for PDB ID 3bfo follow the guidance in the SLIP paper~\citep{thomas2022tuned}. 
    \item For the ML oracle setting, we evaluate our framework on two DMS datasets: Green Fluorescent Proteins (GFP) \citep{sarkisyan2016local} and Adeno-Associated Virus (AAV) \citep{bryant2021deep}. These DMS experiments include up to 15 mutations from the wild-type sequence. The fitness metric for GFP is based on its fluorescence properties as a biomarker, while for AAV, it is based by its ability to package a DNA payload for gene delivery. An ML oracle model is trained following \cite{kirjner2023improving} to predict fitness for any variant.
\end{itemize}

\begin{table}[t]
\centering
\small
\begin{tabular}{lccccc}
\toprule
\multirow{3}{*}{Dataset} & \multirow{3}{*}{Method} & \multirow{3}{*}{Population $\times$ iteration} & \multicolumn{3}{c}{Fitness score} \\ \cmidrule{4-6}
                         &                        &                                  & Top 1 & Top 10 & Top 50 \\ \midrule
                        \multirow{6}{*}{GB1}     & \multirow{3}{*}{EA}    & 32$\times$4    & \textbf{5.38$\pm$1.77}    & \textbf{3.81$\pm$1.10}    & \textbf{2.31$\pm$0.71}    \\ &    
                        & 48$\times$4    & \textbf{4.88$\pm$0.33}    & 3.72$\pm$0.38    & 2.17$\pm$0.27    \\ &    
                        & 96$\times$4    & \textbf{5.72$\pm$0.56}    & \textbf{4.32$\pm$0.53}    & 2.84$\pm$0.60    \\
                        \cmidrule{2-6}
                         & \multirow{3}{*}{Ours} 
                         & 32$\times$4    & 4.34$\pm$0.53    & 3.22$\pm$0.23    & 1.94$\pm$0.28    \\ &    
& 48$\times$4    & 4.31$\pm$0.82    & \textbf{3.76$\pm$0.82}    & \textbf{2.45$\pm$0.61}    \\ &    
& 96$\times$4    & 4.80$\pm$0.52    & 4.09$\pm$0.19    & \textbf{3.04$\pm$0.19}    \\
                        \midrule
                        \multirow{6}{*}{TrpB}    & \multirow{3}{*}{EA}    & 32$\times$4    & 0.20$\pm$0.18    & 0.14$\pm$0.12    & 0.07$\pm$0.05    \\ &    
                        & 48$\times$4    & 0.67$\pm$0.14    & 0.52$\pm$0.11    & 0.19$\pm$0.04    \\ &    
                        & 96$\times$4    & 0.74$\pm$0.01    & 0.59$\pm$0.03    & 0.35$\pm$0.10    \\
                        \cmidrule{2-6}
                         & \multirow{3}{*}{Ours} & 32$\times$4    & \textbf{0.60$\pm$0.10}    & \textbf{0.50$\pm$0.07}    & \textbf{0.35$\pm$0.07}    \\ &    
& 48$\times$4    & \textbf{0.68$\pm$0.04}    & \textbf{0.58$\pm$0.01}    & \textbf{0.36$\pm$0.01}    \\ &    
& 96$\times$4    & \textbf{0.78$\pm$0.20}    & \textbf{0.60$\pm$0.16}    & \textbf{0.39$\pm$0.16}    \\ 

                        \midrule
                        \multirow{6}{*}{Syn-3bfo}   & \multirow{3}{*}{EA}    &  32$\times$8    & 0.57$\pm$0.21    & -0.44$\pm$0.11    & -1.35$\pm$0.17    \\ &    
& 48$\times$8    & 1.29$\pm$0.36    & 0.42$\pm$0.24    & -0.63$\pm$0.07    \\ &    
& 96$\times$8    & 1.85$\pm$0.47    & 1.10$\pm$0.28    & 0.07$\pm$0.28    \\   
                        \cmidrule{2-6}
                         & \multirow{3}{*}{Ours} & 32$\times$8    & \textbf{2.51$\pm$0.23}    & \textbf{1.33$\pm$0.14}    & \textbf{0.28$\pm$0.20}    \\ &    
                        & 48$\times$8    & \textbf{2.35$\pm$0.26}    & \textbf{1.36$\pm$0.11}    & \textbf{0.04$\pm$0.09}    \\ &    
                        & 96$\times$8    & \textbf{2.83$\pm$0.20}    & \textbf{2.02$\pm$0.36}    & \textbf{0.96$\pm$0.36}    \\  
                         \midrule
                        \multirow{6}{*}{AAV}   & \multirow{3}{*}{EA}    & 32$\times$8    &  0.42$\pm$0.03 & 0.36$\pm$0.01& 0.32$\pm$0.00\\ &    
& 48$\times$8    & 0.44$\pm$0.00&0.38$\pm$0.01&0.33$\pm$0.00 \\ &    
& 96$\times$8    & 0.44$\pm$0.00&0.40$\pm$0.01 &0.36$\pm$0.00  \\

                        \cmidrule{2-6}
                         & \multirow{3}{*}{Ours} & 32$\times$8    &  \textbf{0.74$\pm$0.00} &\textbf{0.69$\pm$0.02} &\textbf{0.62$\pm$0.03} \\ &    
                        & 48$\times$8   & \textbf{0.75$\pm$0.01} &\textbf{0.71$\pm$0.01} &\textbf{0.64$\pm$0.02} \\ &     
                        & 96$\times$8    & \textbf{0.76$\pm$0.03} &\textbf{0.73$\pm$0.03} &\textbf{0.68$\pm$0.03}  \\ 

                         \midrule
                        \multirow{6}{*}{GFP}   & \multirow{3}{*}{EA}    & 32$\times$8    &0.43$\pm$0.13 &0.21$\pm$0.02&0.12$\pm$0.01    \\ &     
& 48$\times$8    &0.43$\pm$0.14 &0.26$\pm$0.05 &0.12$\pm$0.01 \\ &    
& 96$\times$8    &0.50$\pm$0.11 &0.34$\pm$0.05 &0.18$\pm$0.01 \\

                        \cmidrule{2-6}
                         & \multirow{3}{*}{Ours} & 32$\times$8    &\textbf{0.96$\pm$0.02} &\textbf{0.94$\pm$0.01} &\textbf{0.88$\pm$0.03}      \\ &    
                        & 48$\times$8   &\textbf{0.96$\pm$0.02} &\textbf{0.93$\pm$0.01} &\textbf{0.84$\pm$0.02}      \\ &    
                        & 96$\times$8   &\textbf{0.97$\pm$0.01} &\textbf{0.95$\pm$0.01} &\textbf{0.92$\pm$0.01}\\

                         \bottomrule
\end{tabular}
\vspace{-0.5em}
\caption{Single-objective optimization results for fitness optimization. We record the mean of top-$k$ ranked candidates and report the mean and std over three random seeds. The best score for different population sizes and landscapes is \textbf{bold}. } 
\label{tab:single-all}
\vspace{-1em}
\end{table}

\vspace{-1em}
\subsection{Main Experiment}

We validate our method in four settings to evaluate our method on protein sequence optimization. 

\textbf{Single-objective optimization.}
We conduct single-objective optimization on all five datasets. In this experiment, we follow the traditional directed evolution protocol, setting the number of proposed variants per iteration to 32, 48, and 96. For GB1 and TrpB, the number of iterations is set to 4, while for the other landscapes, the number of iterations is increased to 8 due to the larger number of possible mutation sites. The optimization objective is to maximize the fitness value from the oracle function, as detailed in \Cref{tab:data_stat}. Among five datasets, GB1 and TrpB have more linear fitness landscapes, where finding a favorable amino acid at a position often leads to its presence in the optimal sequence (demonstrated by \Cref{fig:heatmap}). This allows EA to find strong variants early, sometimes outperforming our method. As shown in \Cref{fig:all_single}, EA only outperforms one of three random seeds for GB1, while our framework performs better in the other two. 

Since linear relationships between positions are less likely in more complex landscapes with larger search spaces, we also evaluate our framework on Syn-3bfo, AAV, and GFP, which have more mutation sites and nonlinear fitness landscapes (\Cref{tab:data_stat}). For Syn-3bfo, the initial pool is generated from single mutations of the wild-type protein 3bfo, with fitness values calculated using the SLIP model. For the AAV and GFP datasets, our initial pool setting follows the medium difficulty criteria outlined in \citep{kirjner2023improving}. This involves restricting the fitness range of the initial pool proteins to fall between a certain range and ensuring that the mutational gap from the highest-score protein in the given dataset is greater than 6. The predicted fitness value is normalized by min-max values of dataset. Our method consistently outperforms EA in these datasets (\Cref{tab:single-all}).

\begin{figure}[t]
\centering
\begin{tabular}{cc}
\includegraphics[width=0.45\textwidth]{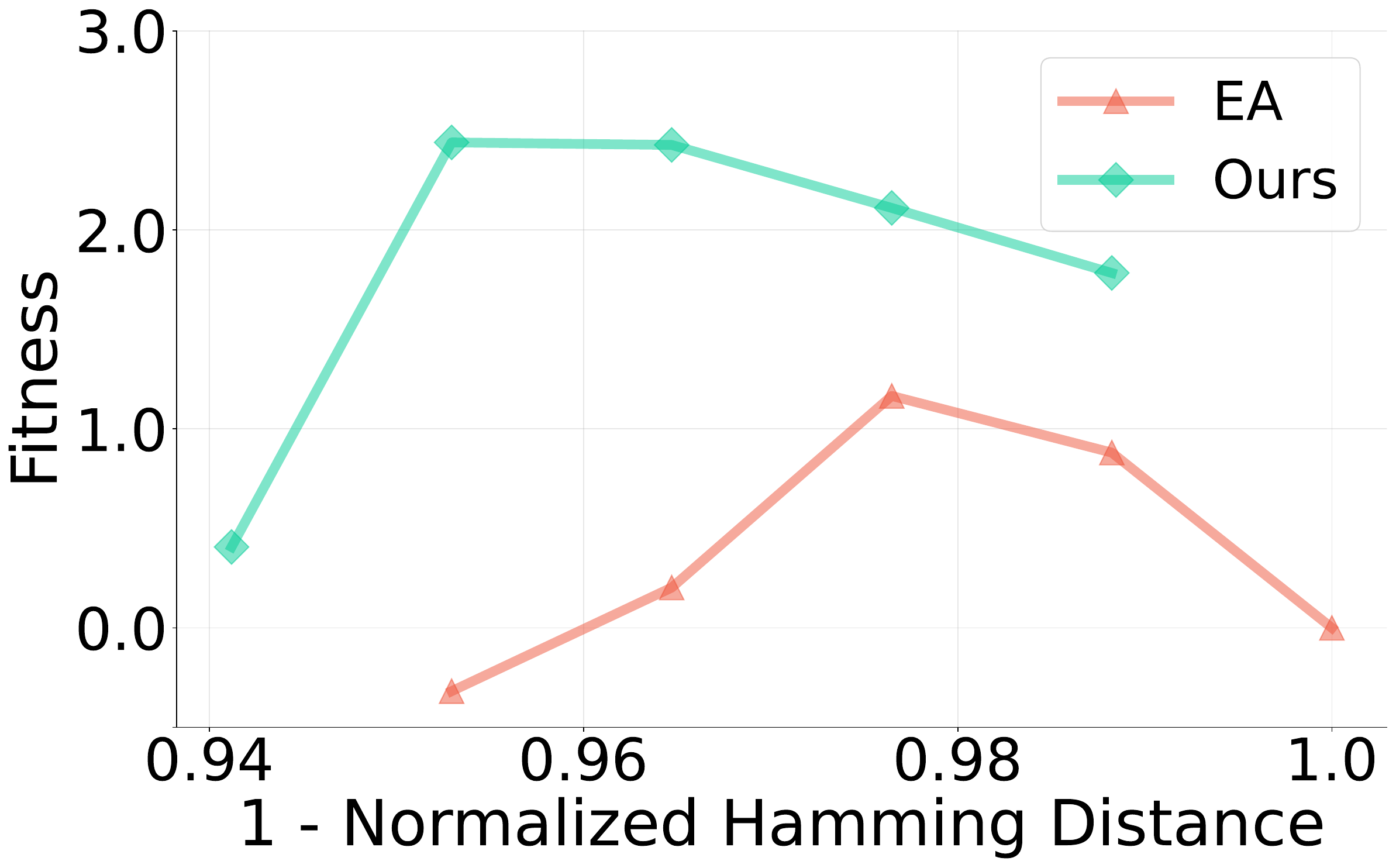} & 
\includegraphics[width=0.45\textwidth]{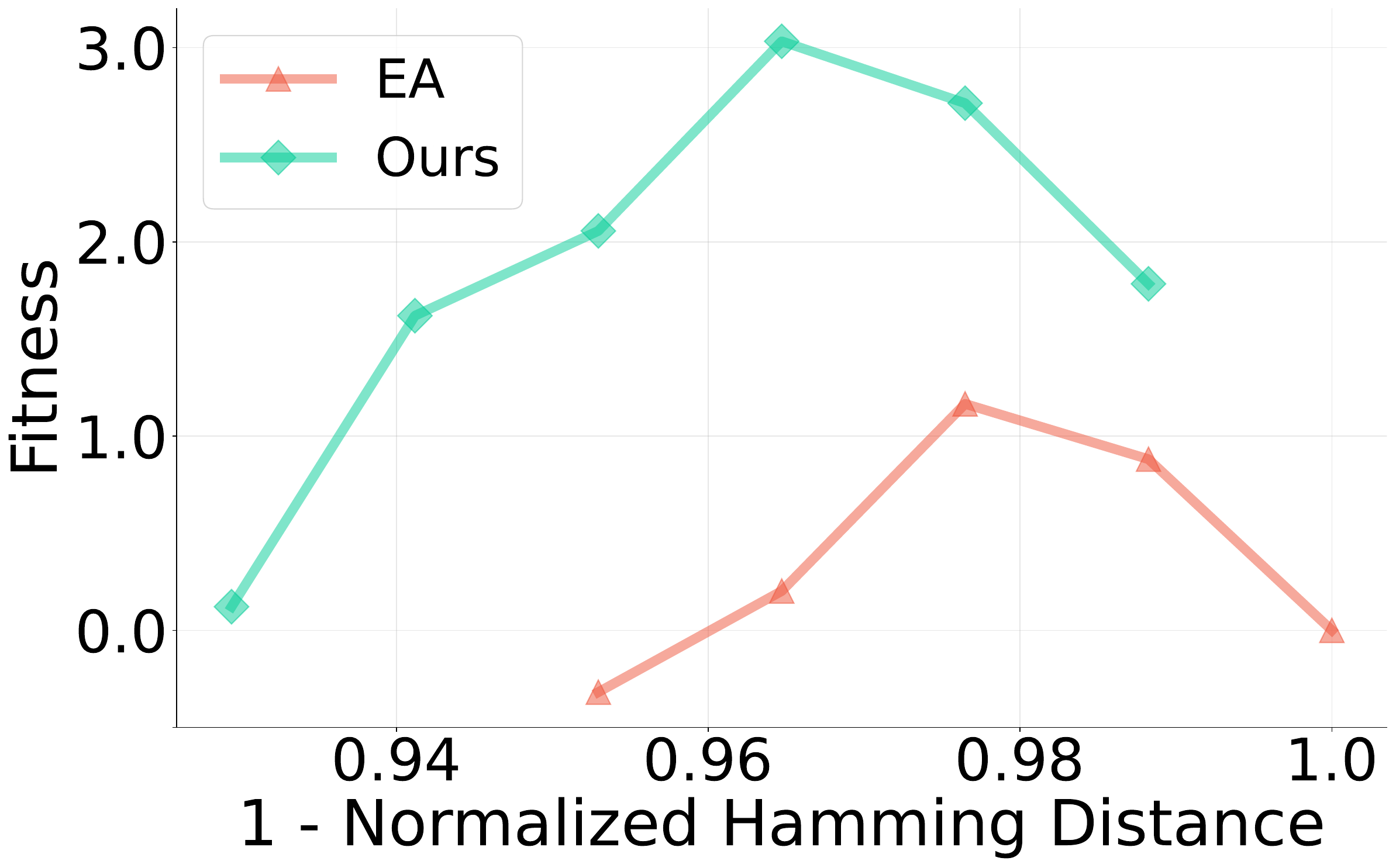}  \\
(a) Constrained optimization & (b) Budget-constrained optimization 
\end{tabular}
\vspace{-0.5em}
\caption{Pareto frontiers identified under constrained and budget-constrained optimization settings.}
\vspace{-0.5em}
\label{fig:constrained_pareto_frontier}
\vspace{-0.5em}
\end{figure}

\begin{figure}[h!]
\centering
\begin{tabular}{cccc}
\!\!\!\!\!\!\includegraphics[width=0.25\textwidth]{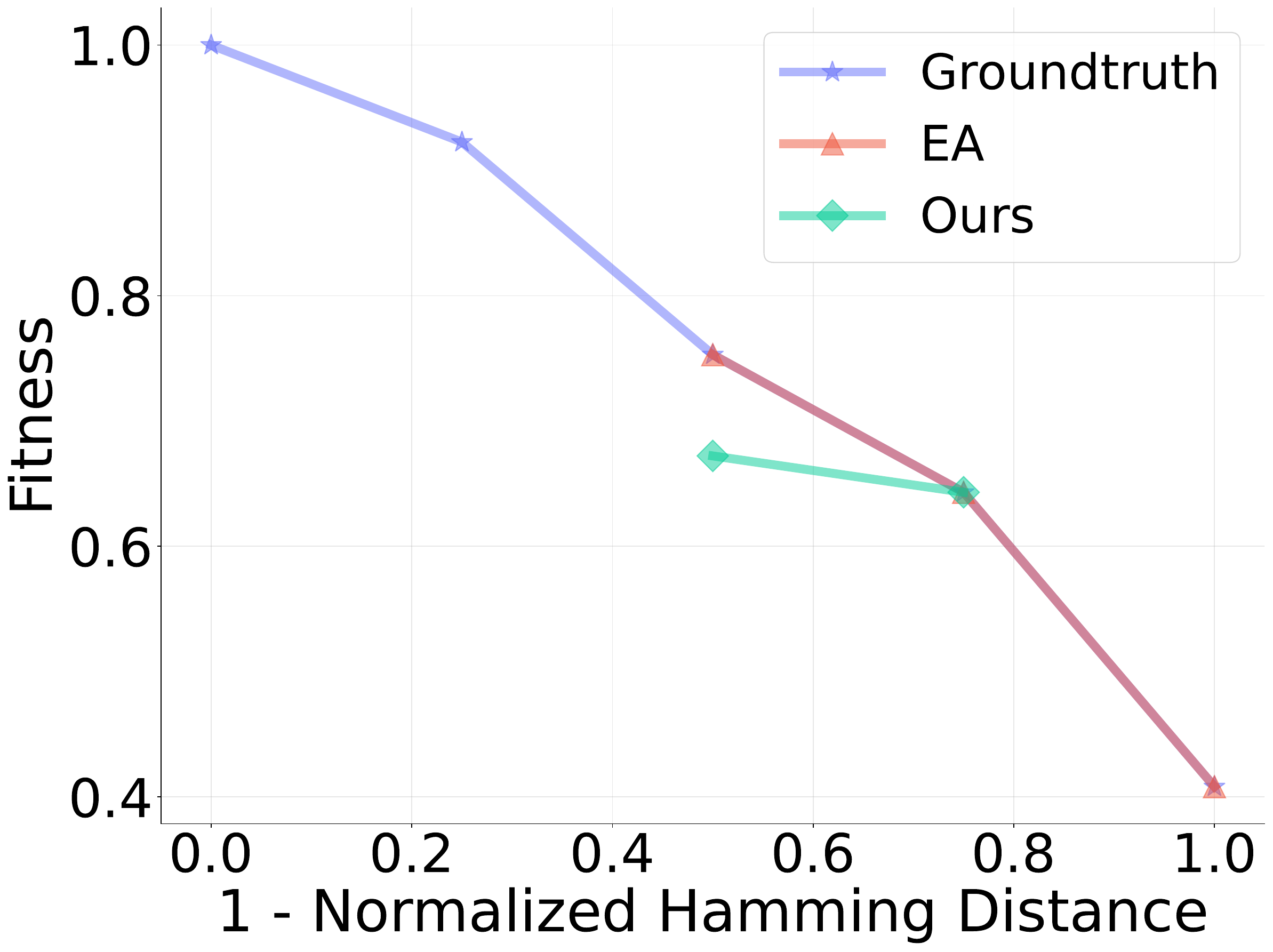}\!\!\!\!&\!\!\!\!
\includegraphics[width=0.25\textwidth]{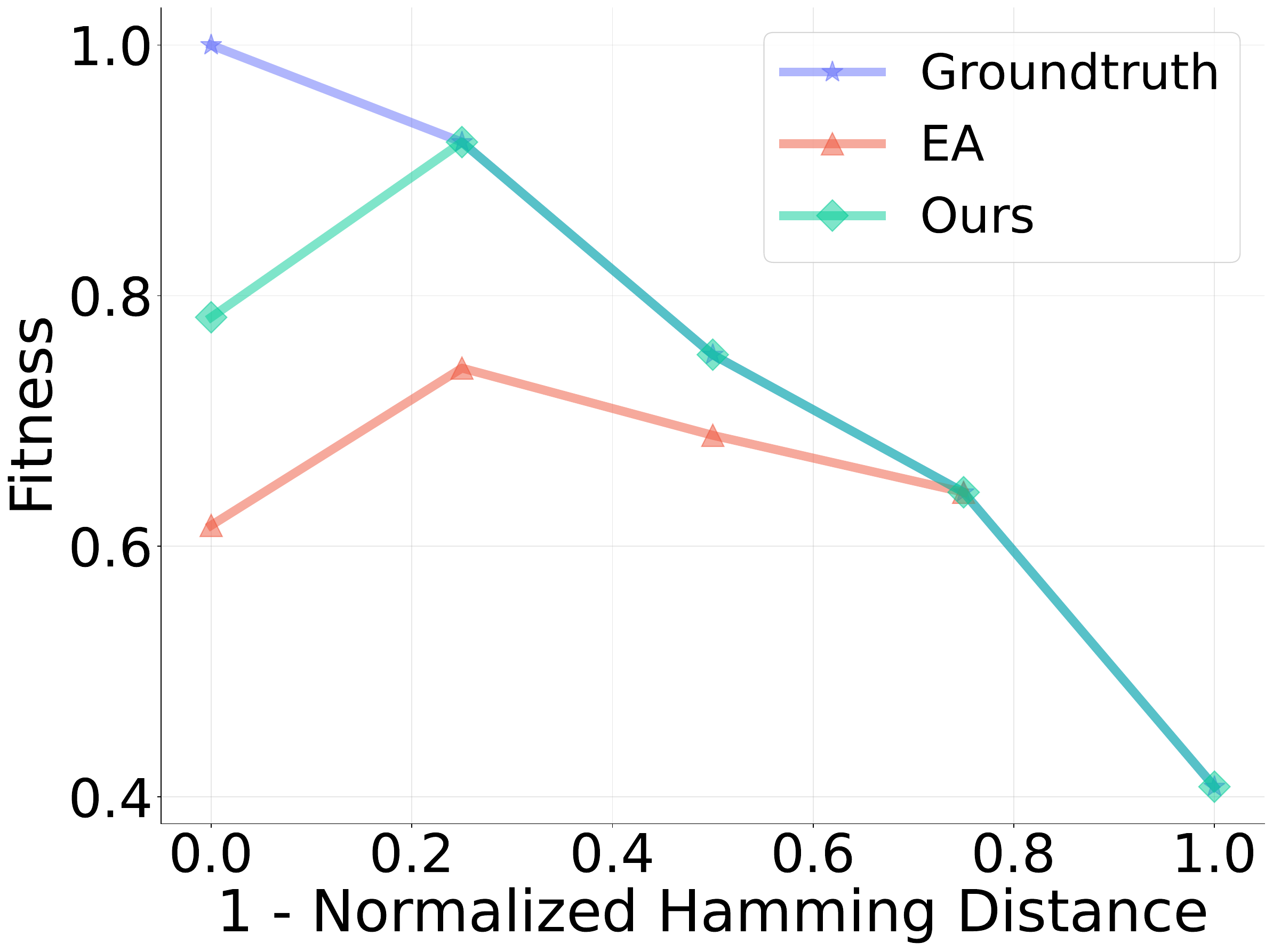}\!\!\!\!&\!\!\!\!
\includegraphics[width=0.25\textwidth]{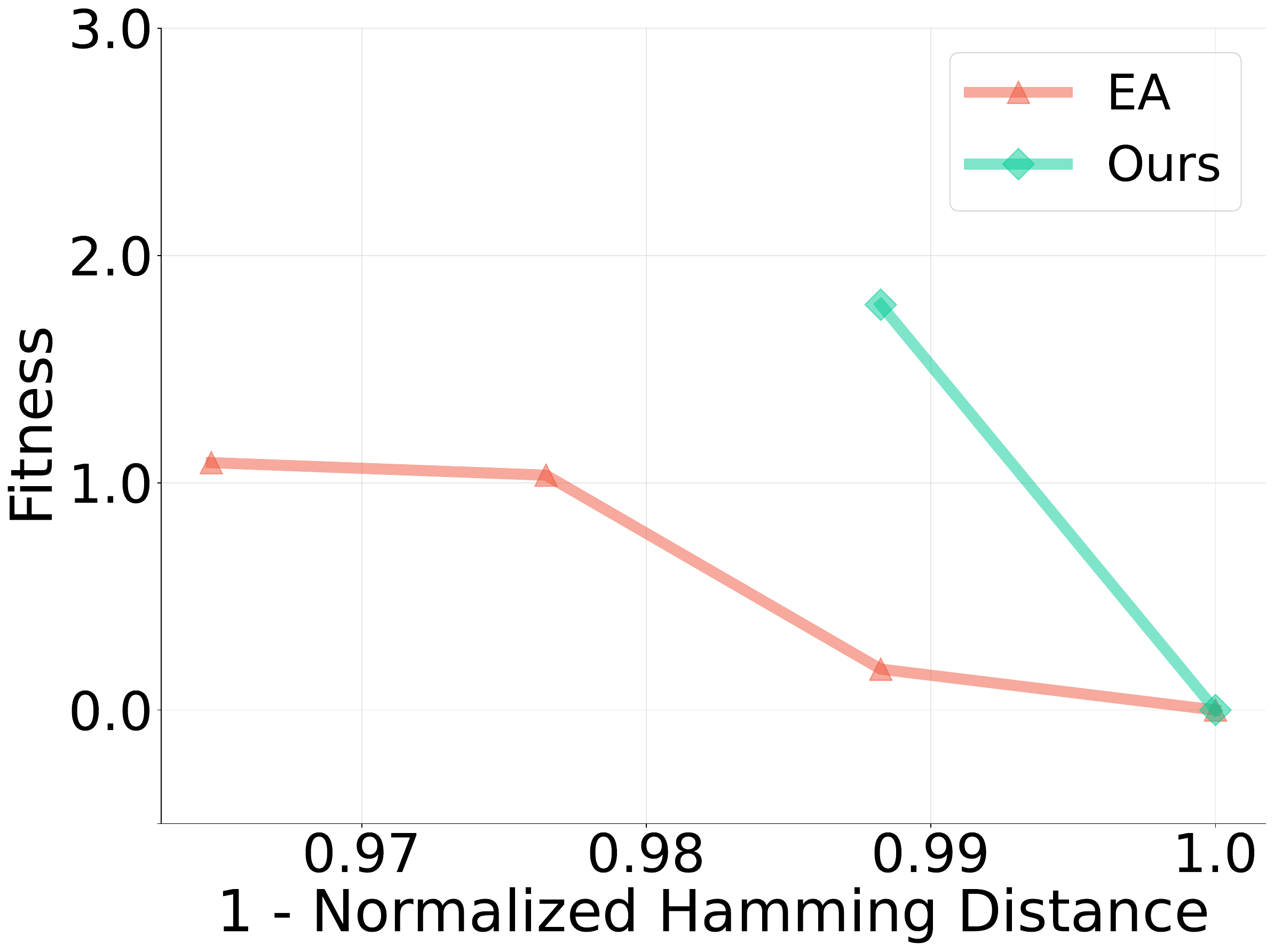}\!\!\!\!&\!\!\!\!
\includegraphics[width=0.25\textwidth]{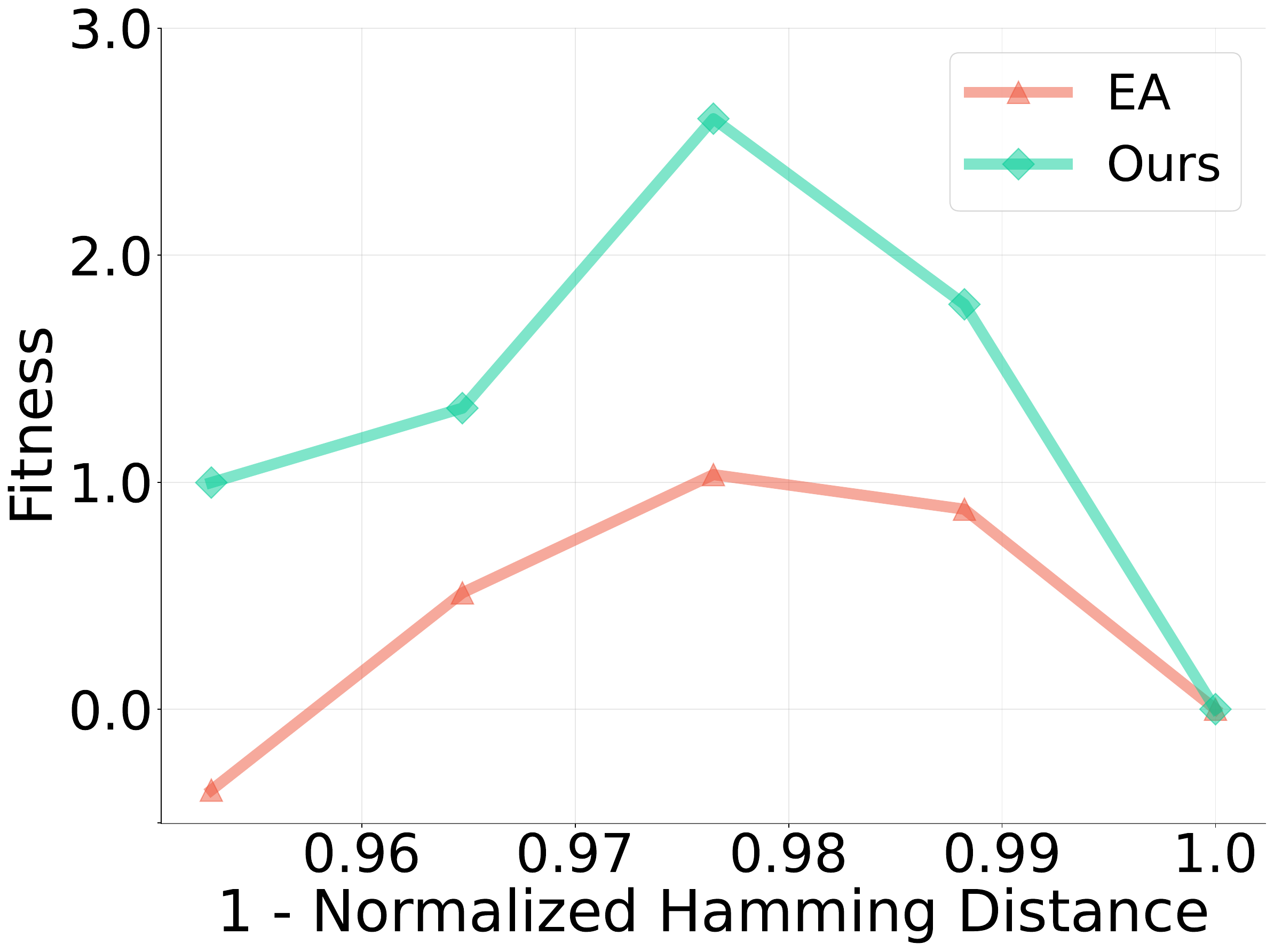}\!\!\!\!\\
\multicolumn{2}{c}{(a) TrpB} & \multicolumn{2}{c}{(b) Syn-3bfo} \\
\end{tabular}
\vspace{-1em}
\caption{Pareto frontiers identified under multi-objective optimizations. We display the Pareto frontiers found for TrpB (a) and Syn-3bfo (b), using \textit{Pareto set selection} (left) and \textit{sum of objectives} (right), respectively. We also show the groundtruth Pareto frontiers for TrpB.}
\label{fig:pareto_frontier}
\vspace{-0.5em}
\end{figure}

\textbf{Constrained optimization.}
In constrained optimization, we limit the number of mutations at each iteration to 3, 5, and 10, rejecting sequences that exceed these limits on the Syn-3bfo landscape. For our method, we add the prompt: \textit{``The proposed sequence must have a Hamming distance between 1 and \{H\} from the \{wild-type sequence\}''}, where \textit{\( H \)} represents the Hamming distance constraint. 

As shown in \Cref{tab:constrained}, our method demonstrates stable performance compared to EA. Notably, the performances of EA at constrained Hamming distances of 5 and 10 are the same, as the maximum Hamming distance of sequences proposed by EA does not exceed 5 within eight iterations. Our framework performs best when the constrained Hamming distance is set to 3. Additionally, we illustrate the Pareto frontier discovered during the constrained optimization tasks by selecting the best fitness value for each Hamming distance from the wild-type in \Cref{fig:constrained_pareto_frontier}.

\textbf{Budget-constrained optimization.} In budget-constrained optimization, we restrict the maximum number of amino acids edited in a single iteration to 1, 2, and 4 on the Syn-3bfo landscape. Sequences exceeding this limit are dropped by rejection sampling. For our method, we include the prompt: \textit{``The proposed sequence must have a Hamming distance between 1 and \{BH\} from the \{parent sequence\}''}, where \textit{\(BH\)} represents the constrained Hamming distance, and \textit{\{parent sequence\}} refers to the two parent sequences provided to the LLM for optimization.

The results in \Cref{tab:budget_constrained} show that our model outperforms EA across all three settings. Our method performs best when the maximum number of amino acids edited in a single iteration is limited to 2. Additionally, we present the Pareto frontier obtained from the budget-constrained optimization task, using the same settings as the constrained optimization shown in \Cref{fig:constrained_pareto_frontier}.

\begin{table*}[t]
\centering
\centering
\scriptsize
\begin{tabular}{l c ccc ccc}
\toprule
\multirow{2}{*}{Dataset}&\multirow{2}{*}{Method}&\multicolumn{3}{c}{{Pareto (Top-$k$)}}&\multicolumn{3}{c}{{Sum (Top-$k$)}} \\
\cmidrule(lr){3-5}\cmidrule(lr){6-8}&&Top1&Top10&Top50&Top1&Top10&Top50 \\
\midrule
\multirow{2}{*}{Syn-3bfo}
&EA&1.38$\pm$1.26&\textbf{0.82$\pm$0.79}&\textbf{0.82$\pm$0.79}
&1.17$\pm$0.27&0.38$\pm$0.22&-0.64$\pm$0.33 \\
&Ours&\textbf{1.52$\pm$0.37}&0.71$\pm$0.26&0.71$\pm$0.26&\textbf{1.73$\pm$0.27}&\textbf{0.82$\pm$0.28}&\textbf{-0.19$\pm$0.37} \\
\midrule
\multirow{2}{*}{TrpB}
&EA&\textbf{0.84$\pm$0.12}&\textbf{0.67$\pm$0.05}&\textbf{0.67$\pm$0.05}&0.67$\pm$0.02&0.60$\pm$0.01&0.47$\pm$0.01\\
&Ours&0.73$\pm$0.14&0.66$\pm$0.08&0.66$\pm$0.08&\textbf{0.69$\pm$0.04}&\textbf{0.62$\pm$0.05}&\textbf{0.50$\pm$0.06}\\
\bottomrule
\end{tabular}
\vspace{-0.6em}
\caption{Multi-objective optimization (Pareto and sum of objectives) with parameter 48$\times$8.}
\label{tab:multi_object}

\vspace{1em}
\centering
\resizebox{\textwidth}{!}{
\begin{tabular}{l c ccc ccc ccc}
\toprule
\multirow{2}{*}{Dataset}&\multirow{2}{*}{Method}
&\multicolumn{3}{c}{H=3}&\multicolumn{3}{c}{H=5}&\multicolumn{3}{c}{H=10} \\
\cmidrule(lr){3-5}\cmidrule(lr){6-8}\cmidrule(lr){9-11}
&&Top1&Top10&Top50&Top1&Top10&Top50&Top1&Top10&Top50 \\
\midrule
\multirow{2}{*}{Syn-3bfo}\!\!\!&\!\!\!EA\!\!\!&\!\!\!1.20$\pm$0.42\!\!\!&\!\!\!0.51$\pm$0.26\!\!\!&\!\!\!-0.57$\pm$0.11\!\!\!&\!\!\!1.29$\pm$0.36\!\!\!&\!\!\!0.42$\pm$0.24\!\!\!&\!\!\!-0.63$\pm$0.07\!\!\!&\!\!\!1.29$\pm$0.36\!\!\!&\!\!\!0.42$\pm$0.24\!\!\!&\!\!\!-0.63$\pm$0.07 \\
\!\!\!&\!\!\!Ours\!\!\!&\!\!\!\textbf{2.46$\pm$0.22}\!\!\!&\!\!\!\textbf{1.74$\pm$0.16}\!\!\!&\!\!\!\textbf{0.66$\pm$0.10}\!\!\!&\!\!\!\textbf{2.21$\pm$0.19}\!\!\!&\!\!\!\textbf{1.59$\pm$0.35}\!\!\!&\!\!\!\textbf{0.49$\pm$0.44}\!\!\!&\!\!\!\textbf{2.28$\pm$0.29}\!\!\!&\!\!\!\textbf{1.74$\pm$0.33}\!\!\!&\!\!\!\textbf{0.73$\pm$0.40}\\
\bottomrule
\end{tabular}}
\vspace{-0.6em}
\caption{Constrained optimization results on Syn-3bfo. Each set of columns shows a different H.}
\label{tab:constrained}

\vspace{1em}
\centering
\resizebox{\textwidth}{!}{
\begin{tabular}{l c ccc ccc ccc}
\toprule
\multirow{2}{*}{Dataset}&\multirow{2}{*}{Method}
&\multicolumn{3}{c}{H=1}
&\multicolumn{3}{c}{H=2}
&\multicolumn{3}{c}{H=4} \\
\cmidrule(lr){3-5}\cmidrule(lr){6-8}\cmidrule(lr){9-11}
&& Top1&Top10&Top50
  &Top1&Top10&Top50
  &Top1&Top10&Top50 \\
\midrule
\multirow{2}{*}{Syn-3bfo}
\!\!\!&\!\!\!EA\!\!\!&\!\!\!1.36$\pm$0.45\!\!\!&\!\!\!0.78$\pm$0.33\!\!\!&\!\!\!-0.24$\pm$0.17\!\!\!&\!\!\!1.42$\pm$0.54\!\!\!&\!\!\!0.62$\pm$0.52\!\!\!&\!\!\!-0.48$\pm$0.35\!\!\!&\!\!\!1.29$\pm$0.36\!\!\!&\!\!\!0.42$\pm$0.24\!\!\!&\!\!\!-0.63$\pm$0.07\\
\!\!\!&\!\!\!Ours\!\!\!&\!\!\!\textbf{2.10$\pm$0.09}\!\!\!&\!\!\!\textbf{1.29$\pm$0.17}\!\!\!&\!\!\!\textbf{0.28$\pm$0.26}\!\!\!&\!\!\!\textbf{2.52$\pm$0.53}\!\!\!&\!\!\!\textbf{1.61$\pm$0.41}\!\!\!&\!\!\!\textbf{0.46$\pm$0.42}\!\!\!&\!\!\!\textbf{2.34$\pm$0.26}\!\!\!&\!\!\!\textbf{1.28$\pm$0.16}\!\!\!&\!\!\!\textbf{-0.01$\pm$0.07} \\
\bottomrule
\end{tabular}}
\vspace{-0.6em}
\caption{Budget-constrained optimization on Syn-3bfo with different budget H.}
\vspace{-1em}
\label{tab:budget_constrained}
\vspace{-1em}

\end{table*}

\textbf{Multi-objective optimization.} We perform multi-objective optimization to simultaneously optimize the Hamming distance and fitness on the Syn-3bfo landscape. In the \textit{sum of objectives} approach, the fitness value and \(1 - \text{normalized Hamming distance}\) are combined into a single objective with equal weight. In the \textit{Pareto set selection} approach, all dominated points are rejected, and optimization is restricted to points on the Pareto frontier.

The results from the first approach are summarized in \Cref{tab:multi_object} and compared against the evolutionary algorithm (EA). The Pareto frontiers identified by our framework and the EA for both approaches are illustrated in \Cref{fig:pareto_frontier}. For the TrpB landscape, the true Pareto frontier can be determined as it is fully enumerated, and our method identifies more Pareto frontier points than EA in the sum-of-objectives setting. Moreover, the Pareto frontiers found by our method in the sum-of-objectives task dominate or are equivalent to those found by EA on both landscapes.

In the Pareto set selection setting, our method does not dominate all the Pareto frontiers identified by EA. This is because restricting the experiment pool to only include Pareto frontier points limits the LLM's access to sufficient information about the sequence space for optimization. However, our method still identifies Pareto frontier points that dominate those found by EA on Syn-3bfo landscape.

\vspace{-0.5em}
\section{Conclusion}
\vspace{-0.5em}
In this paper, we introduce an LLM-guided directed evolution framework for protein sequence optimization. We investigate a range of tasks, employing oracle functions of varying complexity—from synthetic landscapes to experimental ground-truth measurements and machine learning–based oracles. We conduct experiments on multiple optimization tasks from single-objective to constrained and multi-objective optimization. Our results consistently demonstrate the efficacy of LLMs in proposing high-fitness variants. Moving forward, integrating LLM-based optimization into real-world experimental pipelines can accelerate directed evolution experiments, allowing for more efficient exploration of the protein sequence space.

\subsubsection*{Acknowledgments}
We thank Jason Yang for helpful discussions. This work was sponsored by Army Research Office, MURI program, contract \#W911NF2210239.

\newpage
\bibliography{iclr2025_conference}
\bibliographystyle{iclr2025_conference}

\newpage
\appendix
\section{Appendix}

\subsection{Pseudocode}
We show the pseudocode of our framework below.

\begin{algorithm}[h]
\caption{Protein Sequence Optimization with LLM}\label{alg:protein_opt}
\KwData{Initial population $\mathcal{P}_0$; mutation rate $r_m$; population size $K$; number of iterations $N$; the fitness function $F(\cdot)$; the default crossover function $C(\cdot, \cdot)$; the default mutation function $M(\cdot)$.}
\KwResult{Optimized protein population $\mathcal{P}_N$.}
\Begin{
    \For{$s \in \mathcal{P}_0$}{
        Compute $F(s)$\;
    }
    \For{$t \in [1, N]$}{
        \texttt{offspring = []}\;
        \For{$k \in [1, K]$}{
            Draw parent sequences $(s_0, s_1) \sim \mathcal{P}_t \times \mathcal{P}_t$\;
            \texttt{proposed\_seq} $\leftarrow$ \texttt{LLM\_propose}($s_0, s_1$)\;
            \If{\texttt{proposed\_seq} is \texttt{None}}{
                \texttt{offspring.append(}$C$($s_0, s_1$)\texttt{)}\;
                $r \sim \texttt{Uniform}[0, 1]$
                
                \If{$r \leq r_m$}{
                    \texttt{offspring.append($M$}($s_0$)\texttt{)}\;
                }
            }
            \Else{
                \texttt{offspring.append(proposed\_seq)}\;
            }
        }
        \For{$s \in \texttt{offspring}$}{
            Compute $F(s)$\;
        }
        \texttt{merged\_population} $\leftarrow$ \texttt{merge}($\mathcal{P}_t, \texttt{offspring}$)\;
        $\mathcal{P}_t \leftarrow$ \texttt{sorted}(\texttt{merged\_population})[:$K$]\;
    }
    Return $\mathcal{P}_N$\;
}
\end{algorithm}

\subsection{Datasets analyze}
\begin{figure}[ht]
\centering
\begin{tabular}{cc}
\includegraphics[width=0.45\textwidth]{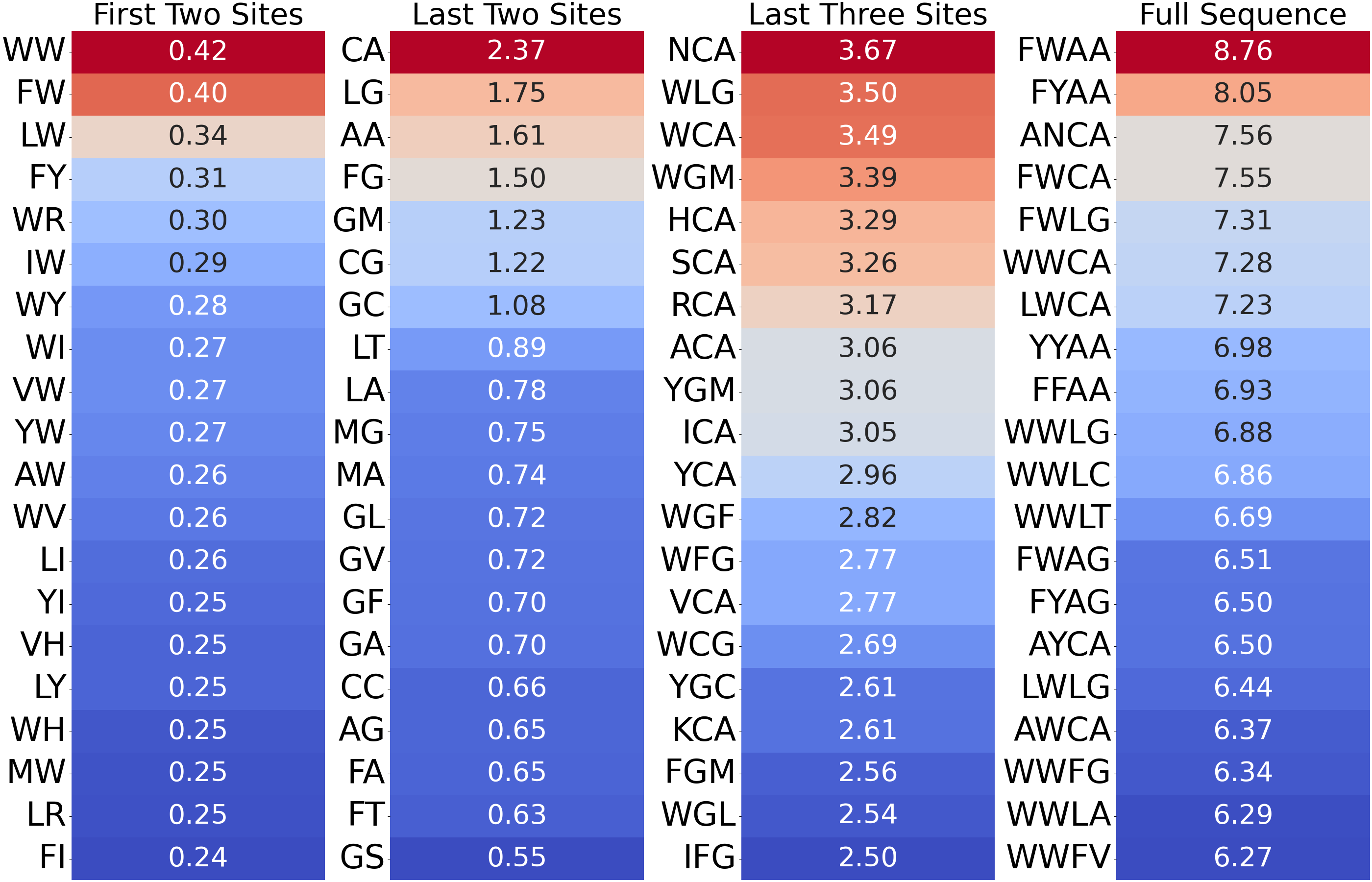}  & \includegraphics[width=0.45\textwidth]{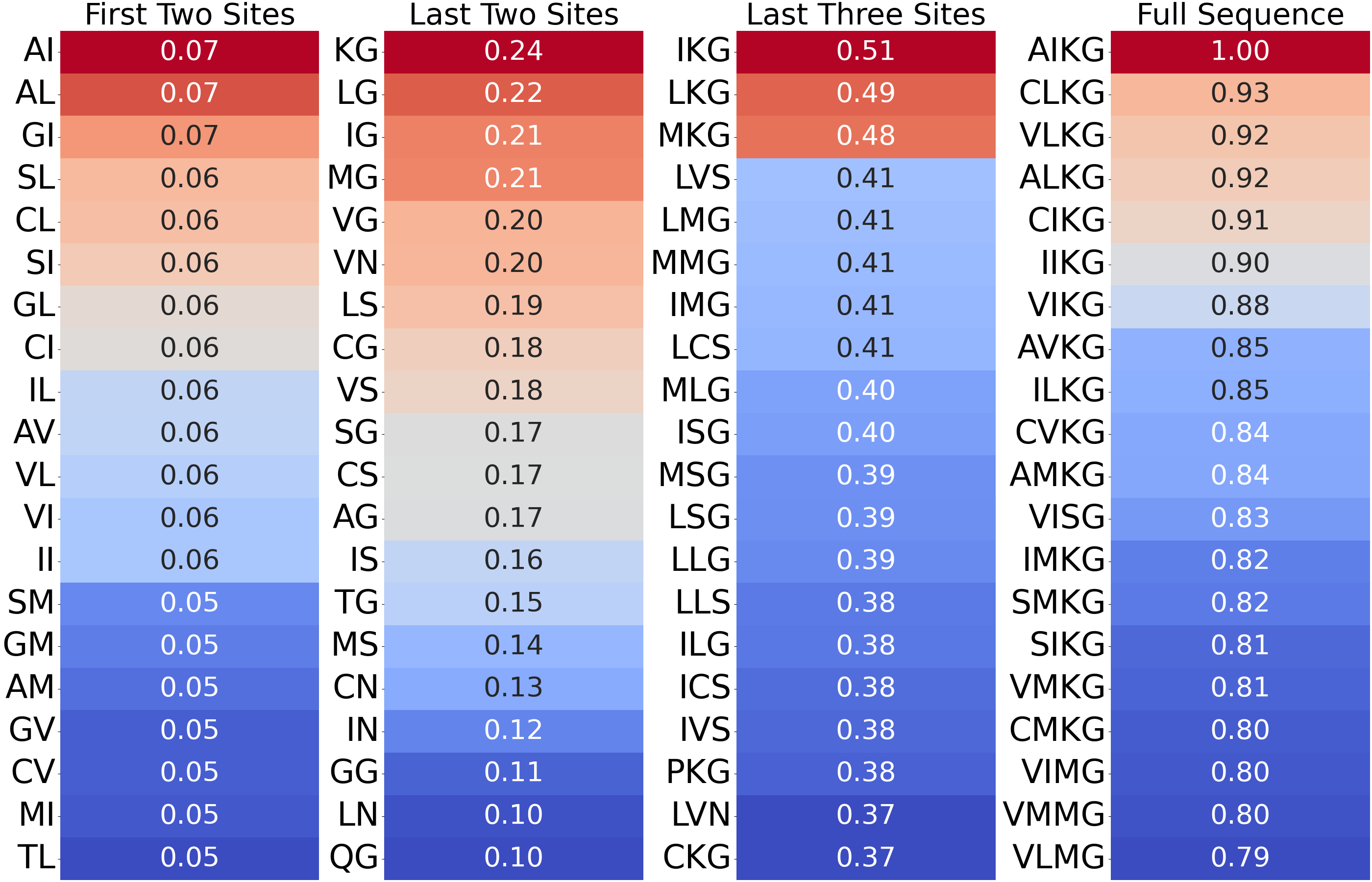} \\
(a) GB1 & (b) TrpB \\
\end{tabular}
\caption{The fitness heatmaps of first two, last two, last three, and full sequence on two datasets.}
\label{fig:heatmap}
\end{figure}

We present a heatmap of the average scores for specific combinations at different positions: the first two, last two, last three, and the full sequence for GB1 and TrpB in \Cref{fig:heatmap}. The heatmap reveals that certain combinations in the last two positions, such as CA, LG, and AA in GB1, and KG, LG, and IG in TrpB, exhibit higher fitness scores compared to others. Preserving these combinations can significantly simplify the path to identifying sequences with improved fitness.

\subsection{Prompts}
The example prompts we use for GB1 is shown below, while the dataset description varies for different datasets. 
\vspace{-1cm}
\begin{tcolorbox}[colback=blue!5!white, colframe=blue!75!black, title=Prompt]
\begin{tcolorbox}[colback=headercolor, colframe=black, sharp corners=all, width=\textwidth, boxrule=0.5mm, title=\texttt{system}]
You are a world-class assistant specializing in protein engineering, fitness optimization, and sequence design. Your expertise lies in analyzing sequence-function relationships, interpreting experimental data, and proposing rational modifications to optimize protein fitness.
\end{tcolorbox}

\vspace{-2mm}

\begin{tcolorbox}[ colback=contentcolor, colframe=black, sharp corners=all, width=\textwidth, boxrule=0.3mm,
title=\texttt{user}]
You will carry out a multi-round directed evolution experiment with the following protein sequence, aimed at improving protein's ability to bind affinity-based sequence enrichment via protein fitness optimization.

\textbf{\#\#\# Protein fitness optimization}

The fitness score reflects the efficacy or functionality for a desired application, from chemical synthesis to bioremediation and therapeutics. Protein fitness optimization can be thought of as navigating a protein fitness landscape, a mapping of amino acid sequences to fitness values, to find higher-fitness variants. Specifically, it is achieved by making crossover and mutations on the given sequences.

We are focusing on changes to a limited subset of amino acids within the sequence. The provided subset protein sequences come from B1 domain of streptococcal protein G, with sequence:

\`{}\`{}\`{}

\begin{minipage}{\textwidth}
\seqsplit{%
MTYKLILNGKTLKGETTTEAVDAATAEKVFKQYANDNGVDGEWTYDDATKTFTVTE}
\end{minipage}

\`{}\`{}\`{}

Each subset protein sequence represents specific amino acid substitutions at four key positions: 39, 40, 41, and 54, denoted using the single-letter amino acid code.

\textbf{\#\#\# Parent protein sequences}

Here are the parent protein sequences that you will be modifying from. Each sequence comes with 4 amino acids and its fitness score is also provided.

Protein sequence 1 (fitness score: 0.0018)

\`{}\`{}\`{}

Q H V R

\`{}\`{}\`{}

Protein sequence 2 (fitness score: 0.0021)

\`{}\`{}\`{}

R L I V

\`{}\`{}\`{}

\textbf{\#\#\# Instructions}

Follow the instructions below to propose a new protein:

* Your proposal should focus on maximizing fitness and minimizing humming distance from the wild type while considering structural and functional plausibility.

* You can propose it via making crossover or mutation on the parent sequences.

* You can also propose a new sequence based on your knowledge.

* Your proposed sequence MUST have the same length as the parent sequences.

* DO NOT propose sequence that is identical with the parent or the wild type sequences.

* Your output MUST ONLY include: \texttt{\textbackslash box\{\{Protein\}\}}.

\end{tcolorbox}

\vspace{1em}
\end{tcolorbox}

\vspace{-1em}

\subsection{Pareto Frontier}
\vspace{-0.5em}

\begin{figure}[!h]
    \begin{tabular}{ccc}
         \!\!\!\!\includegraphics[width=0.33\linewidth]{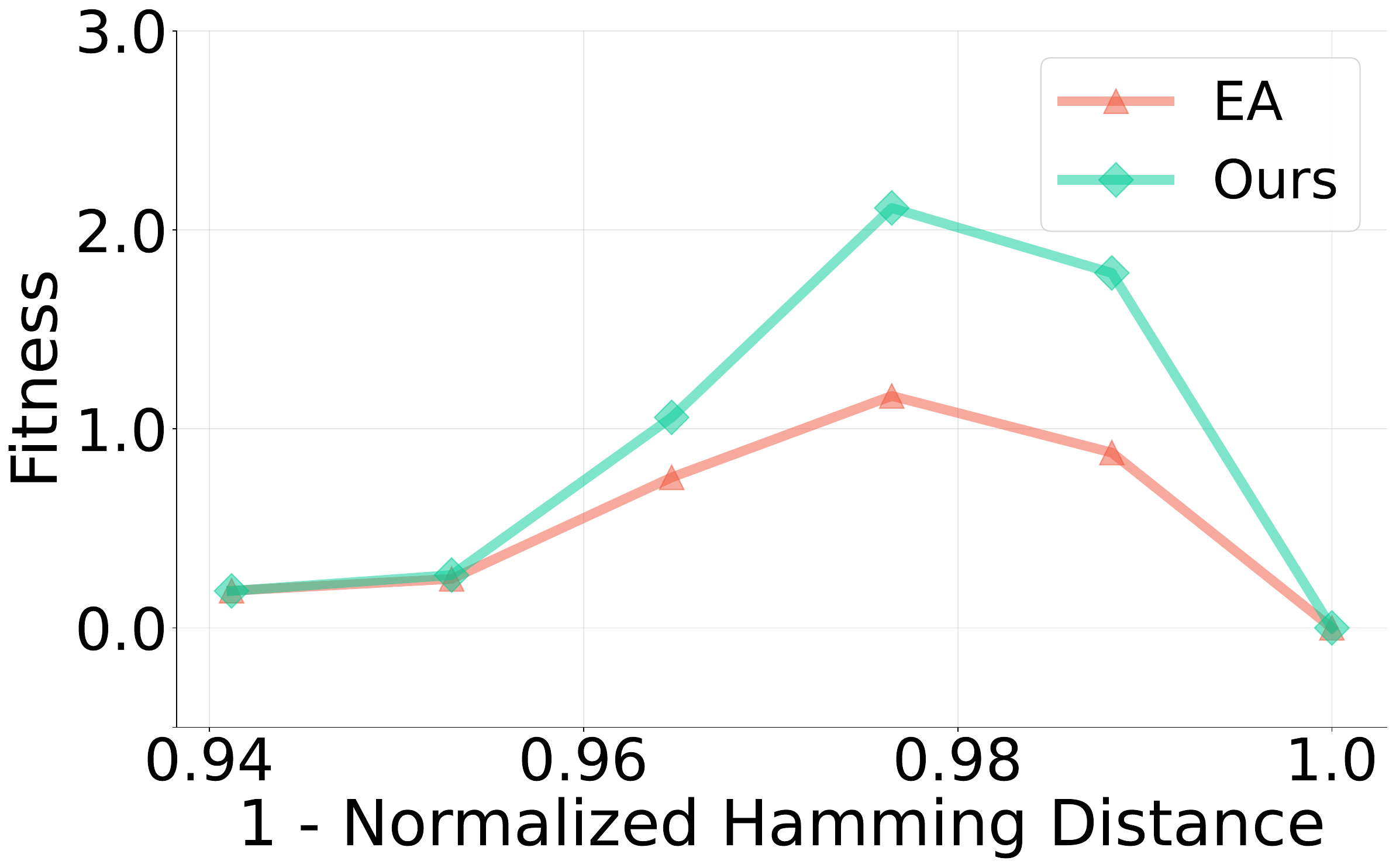}\!\!\!\!&  
         \!\!\!\!\includegraphics[width=0.33\linewidth]{figures/Pareto_Frontier_BK2_Syn-3bfo.pdf}\!\!\!\! & 
         \!\!\!\!\includegraphics[width=0.33\linewidth]{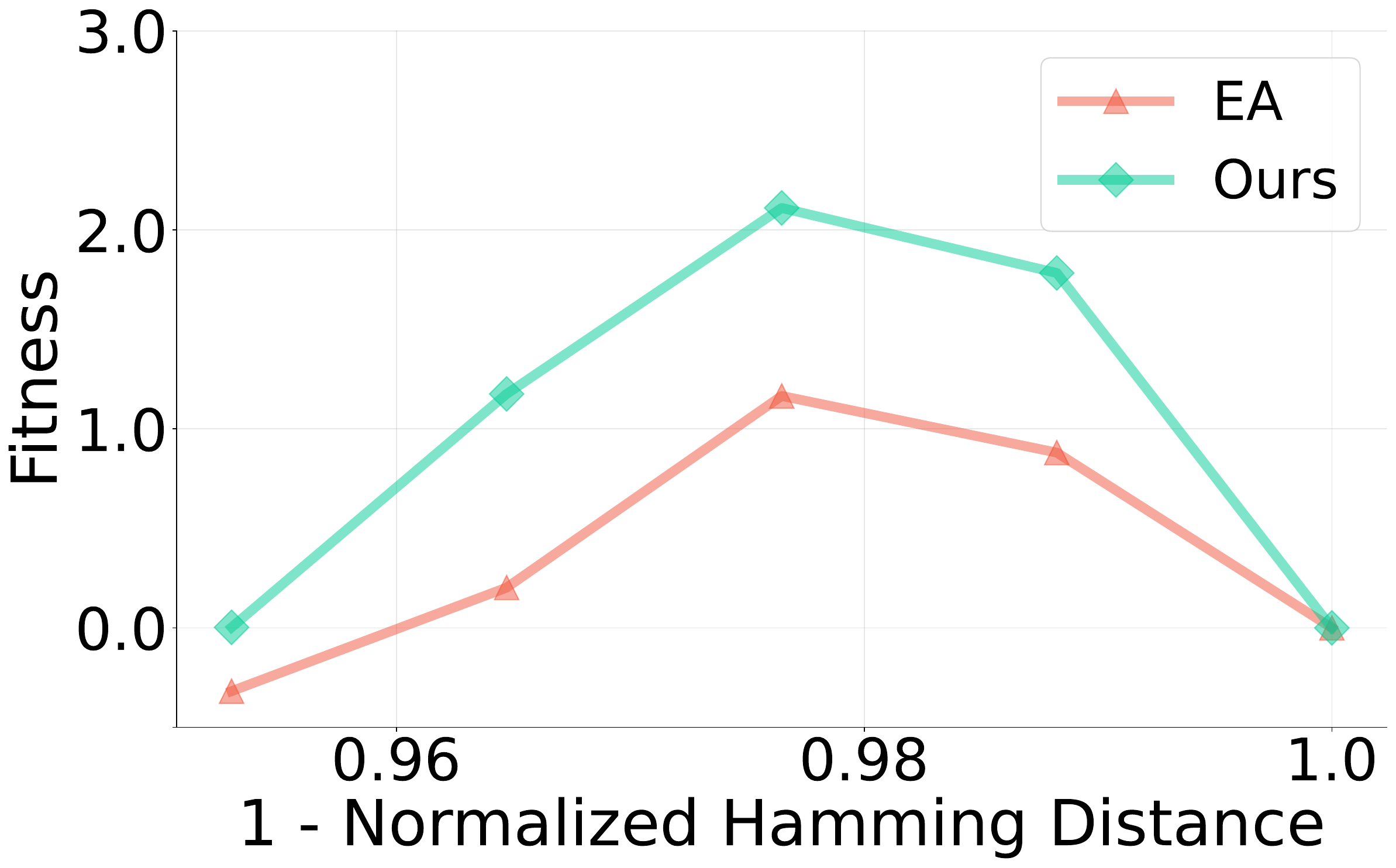}\!\!\!\!\\
        \!\!\!\!Budget \( H = 1 \)\!\!\!\! &\!\!\!\! Budget \( H = 2 \)\!\!\!\! &\!\!\!\! Budget \( H = 4 \)\!\!\!\!\\
        \!\!\!\!\includegraphics[width=0.33\linewidth]{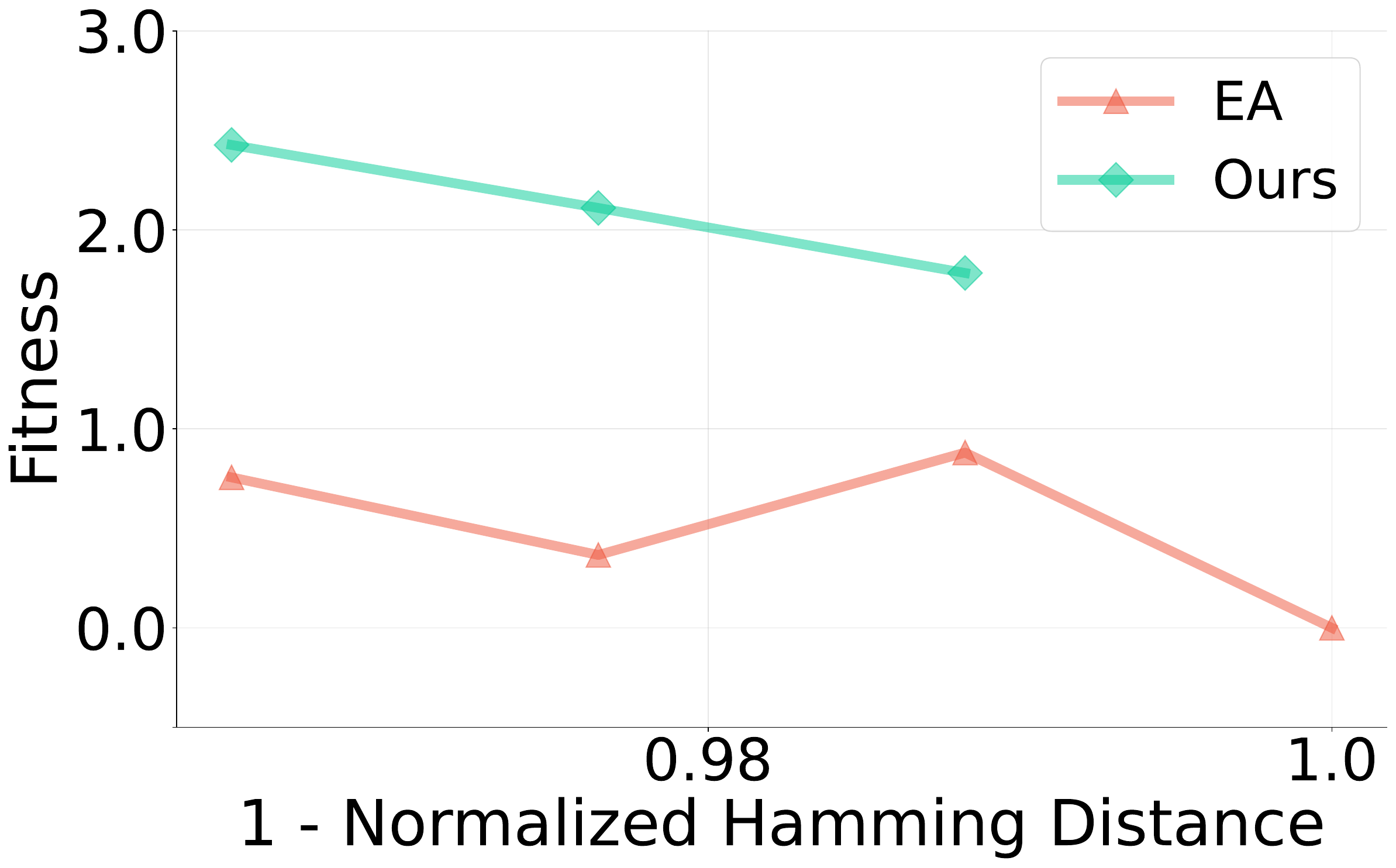}\!\!\!\!&  
        \!\!\!\!\includegraphics[width=0.33\linewidth]{figures/Pareto_Frontier_K5_Syn-3bfo.pdf}\!\!\!\! & 
        \!\!\!\!\includegraphics[width=0.33\linewidth]{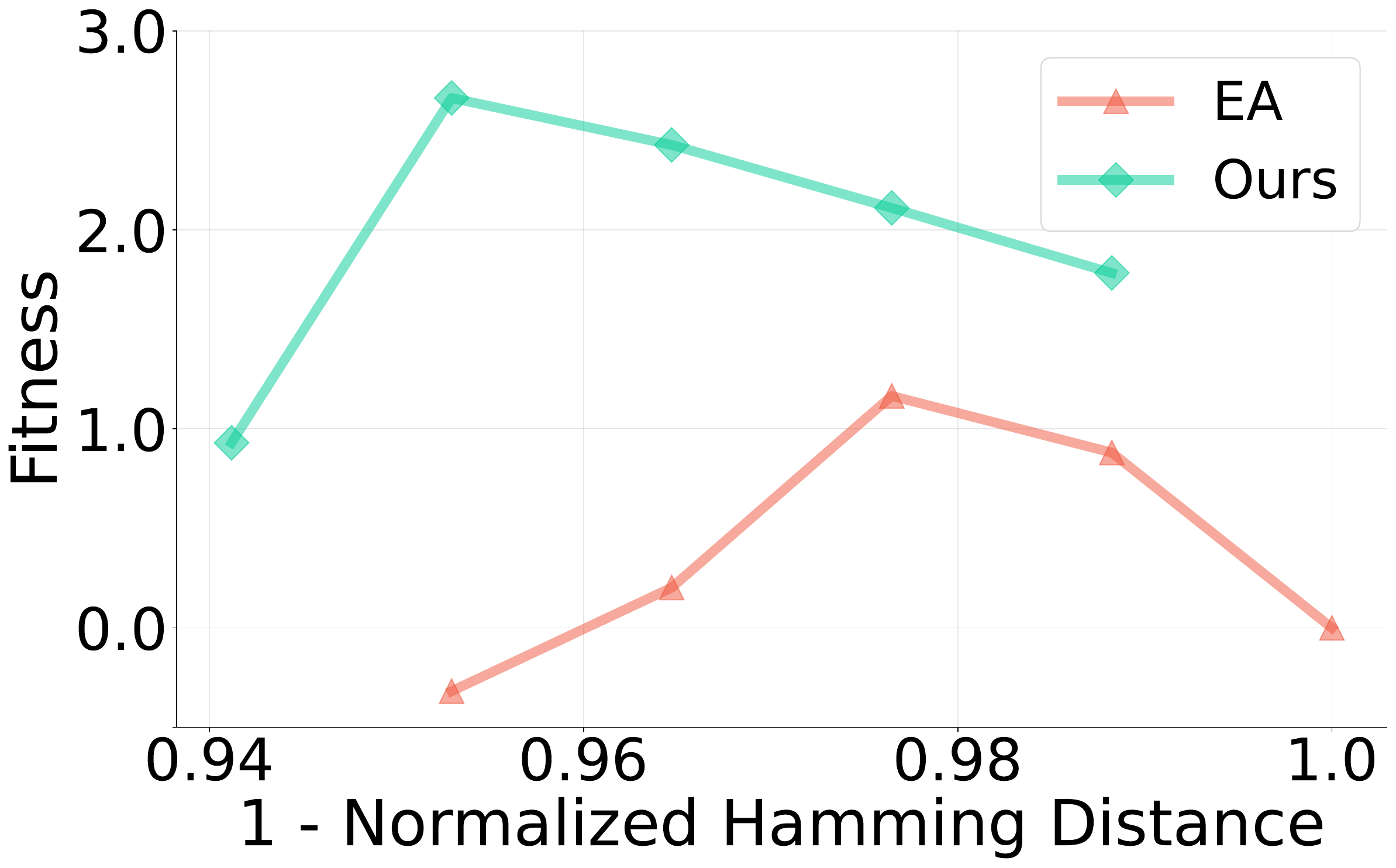}\!\!\!\!\\
        \!\!\!\! \( H = 3 \) \!\!\!\! &\!\!\!\! \( H = 5 \) \!\!\!\! & \!\!\!\! \( H = 10 \) \!\!\!\!\\
    \end{tabular}
    \vspace{-1em}
    \caption{The Pareto frontiers identified by EA and our method in both constrained and budget-constrained optimization settings for all parameter configurations.}
    \vspace{-1em}
    \label{fig:pareto_constrained}
\end{figure}

We present the Pareto frontiers identified through constrained optimization tasks with different Budget \( H \) and \( H \), as shown in \Cref{fig:pareto_constrained}. The figures shows our method consistently dominate EA.

\begin{figure}[ht]
    \centering
    \includegraphics[width=0.5\linewidth]{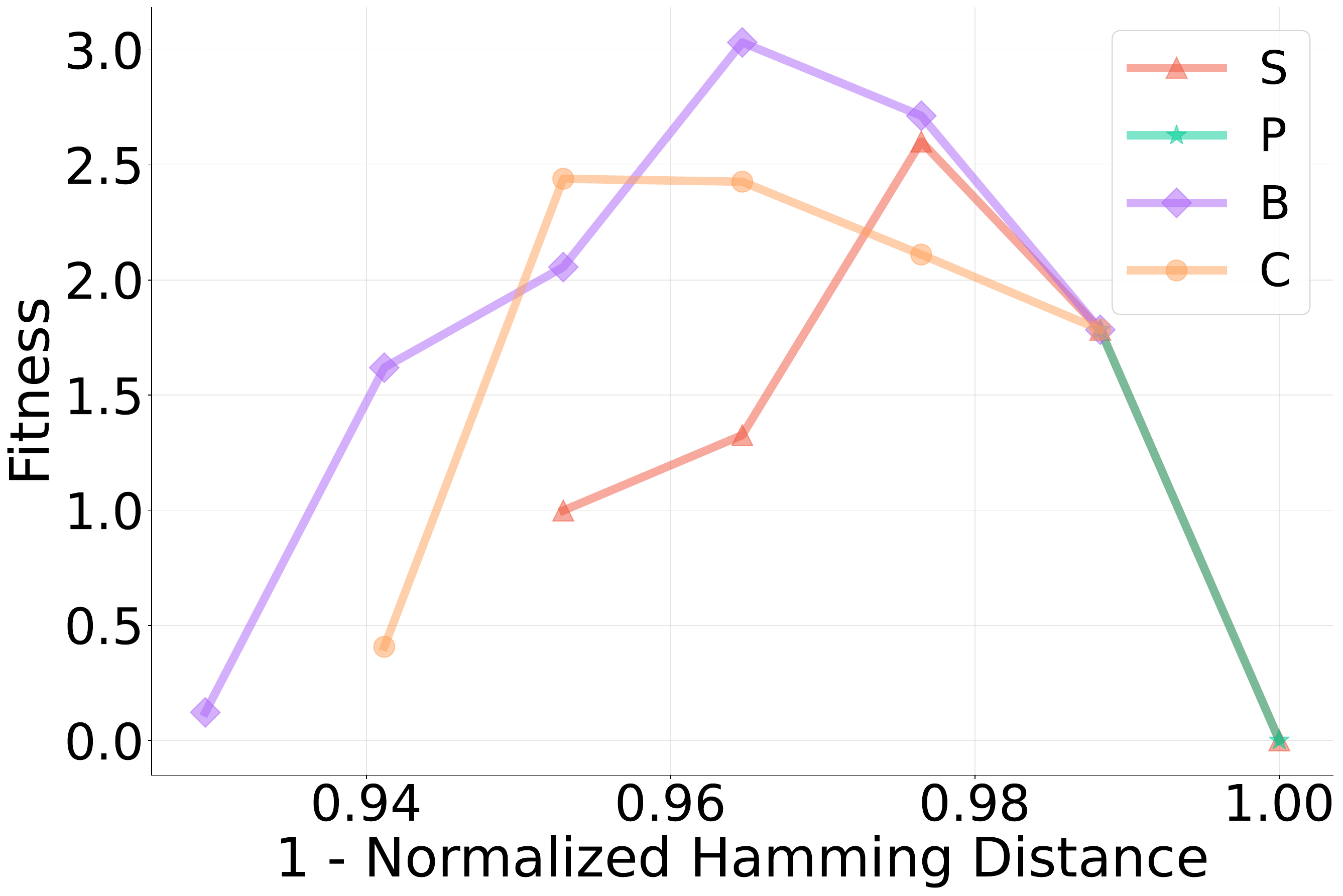}
    \vspace{-1em}
    \caption{The Pareto frontier found by our method via different task. S stand for \textit{Sum of objective}, P stand for \textit{Pareto frontier set selection}, B stand for Budget-constrained, and C stand for the Constrained.}
    \vspace{-1em}
    \label{fig:pareto_all}
\end{figure}

We also present the Pareto frontiers identified by our method for different tasks on the Syn-3bfo dataset in \Cref{fig:pareto_all}, illustrating how the choice of objectives to optimize influences the discovery of the Pareto frontier.
\vspace{1mm}
\vspace{-1em}
\subsection{Ablation Study of the Number of Iterations}
\vspace{-0.5em}
We analyze the impact of varying the number of iterations on the results in \Cref{tab:ablation_study}. The analysis shows that as the number of iterations increases, both EA and our framework improve in performance. However, our method consistently maintains its advantage over EA.

\begin{table}[ht]
\centering
\small
\begin{tabular}{cccccc}
\toprule
\multirow{3}{*}{Dataset} & \multirow{3}{*}{Method} & \multirow{3}{*}{Iteration} & \multicolumn{3}{c}{Fitness score} \\ \cmidrule{4-6}
                         &                        &                                  & Top 1 & Top 10 & Top 50 \\ \midrule
                        \multirow{6}{*}{Syn-3bfo}     & \multirow{3}{*}{EA} &   
                        8    &1.85$\pm$0.47    & 1.10$\pm$0.28    & 0.07$\pm$0.28  \\
                         & & 12& 1.97$\pm$0.58    & 1.52$\pm$0.41    & 0.77$\pm$0.27 \\
                         & & 16 & 2.38$\pm$0.45    & 1.99$\pm$0.42    & 1.34$\pm$0.34 \\
                         \cmidrule{2-6}
                         & \multirow{3}{*}{Ours} & 8    & \textbf{2.83$\pm$0.20}    & \textbf{2.02$\pm$0.36}    & \textbf{0.96$\pm$0.36}   \\
                         & & 12& \textbf{3.03$\pm$0.29}    & \textbf{2.51$\pm$0.36}    & \textbf{1.66$\pm$0.44} \\
                         & & 16 & \textbf{3.84$\pm$0.44}    & \textbf{3.29$\pm$0.36}    & \textbf{2.48$\pm$0.34}\\
                         
                        \bottomrule
\end{tabular}
\vspace{-1em}
\caption{Ablation study on different iterations for Syn-3bfo landscape with 96 population size.} 
\label{tab:ablation_study}
\vspace{-1em}
\end{table}

\vspace{-0.5em}
\subsection{Additional experiments result }
\begin{figure}[ht]
\centering
\begin{tabular}{ccc}
\includegraphics[width=0.3\textwidth]{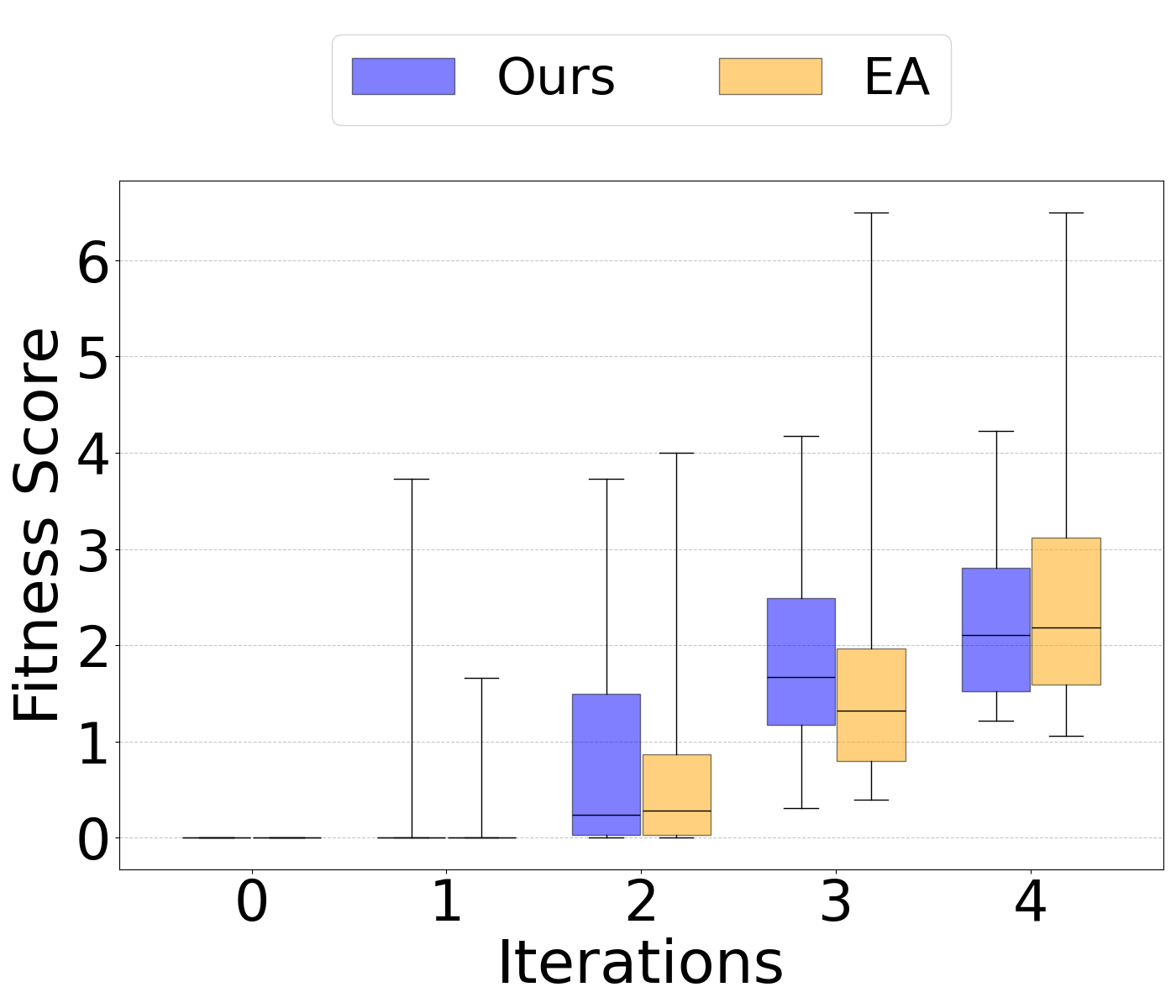} & 
\includegraphics[width=0.3\textwidth]{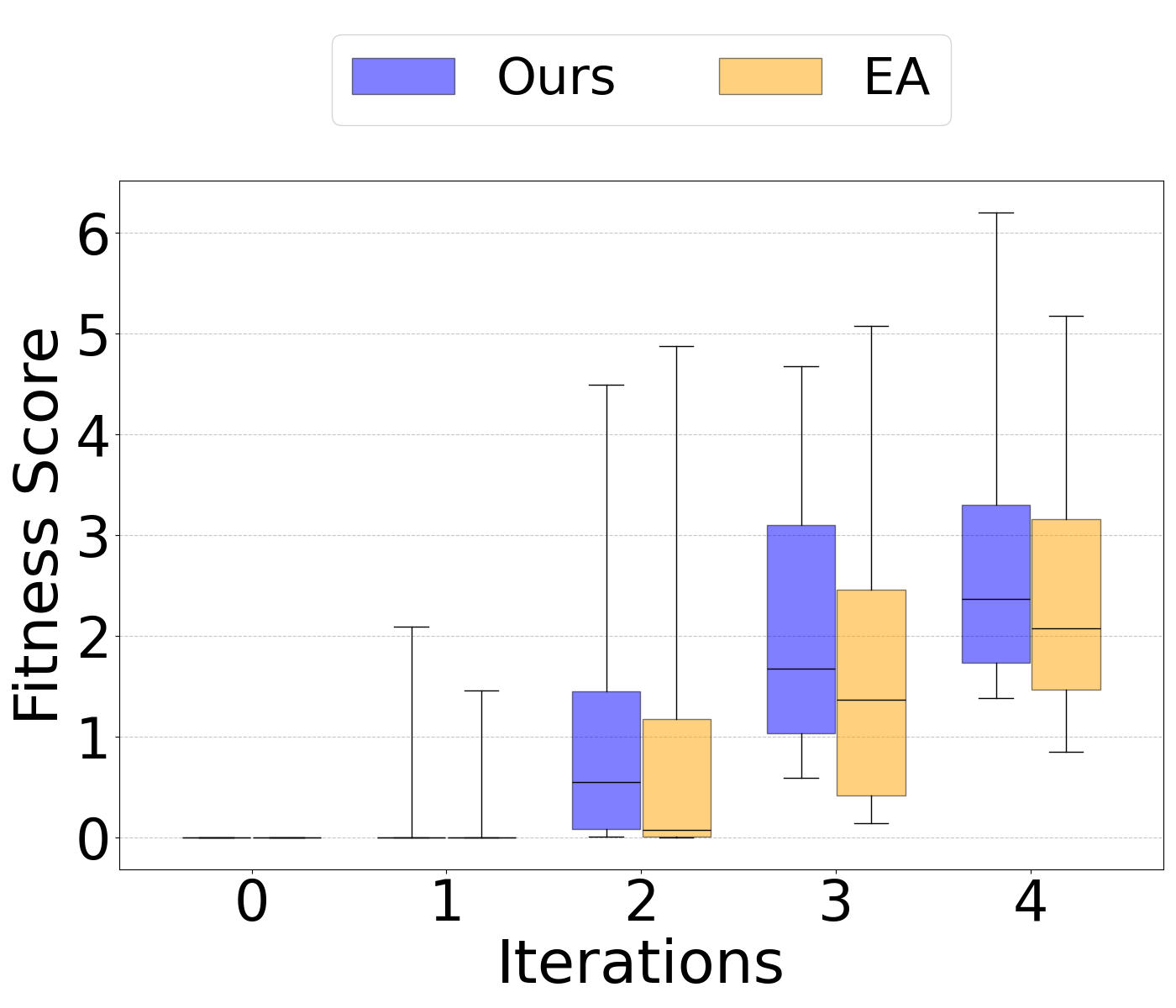} &
\includegraphics[width=0.3\textwidth]{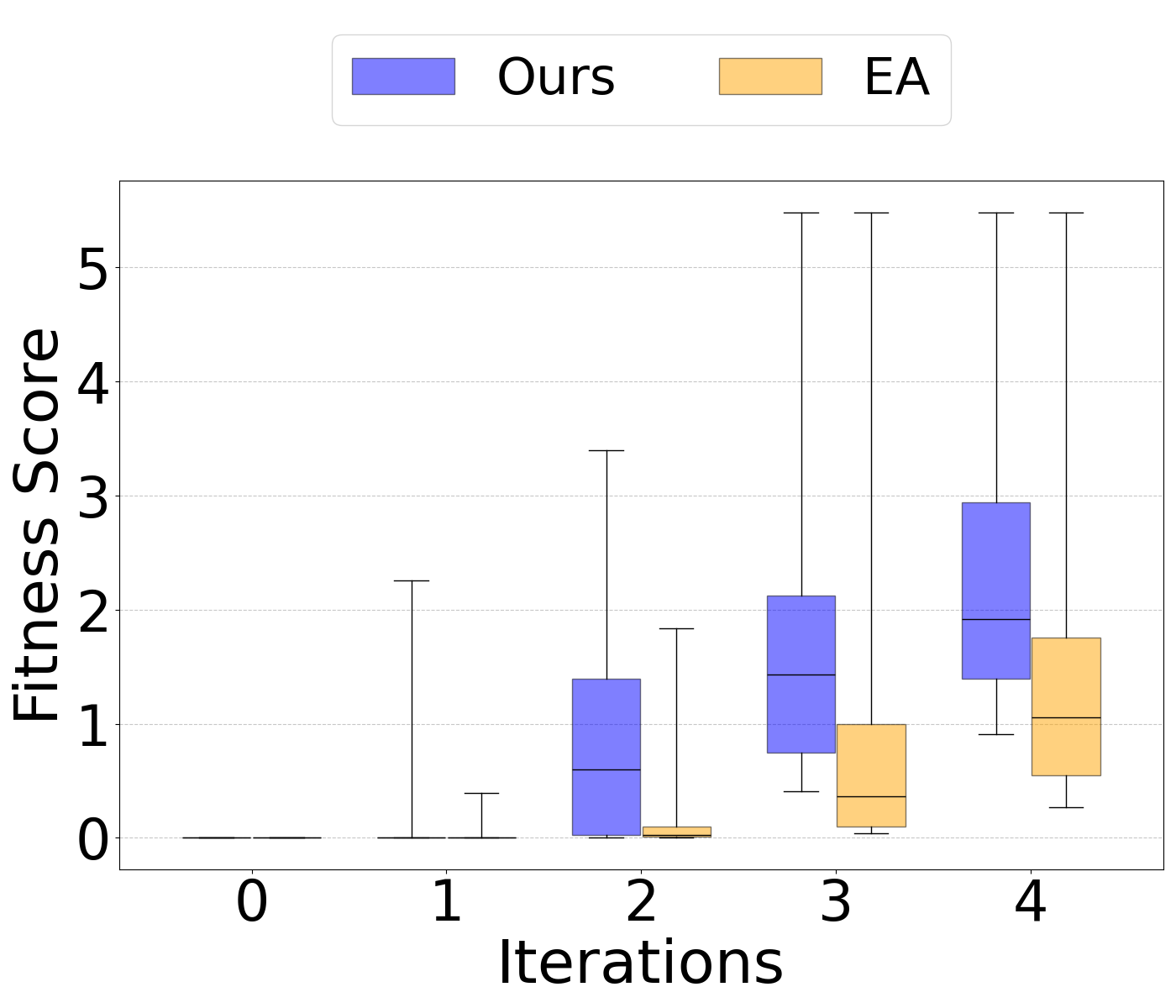} \\
\multicolumn{3}{c}{GB1}  \\
\includegraphics[width=0.3\textwidth]{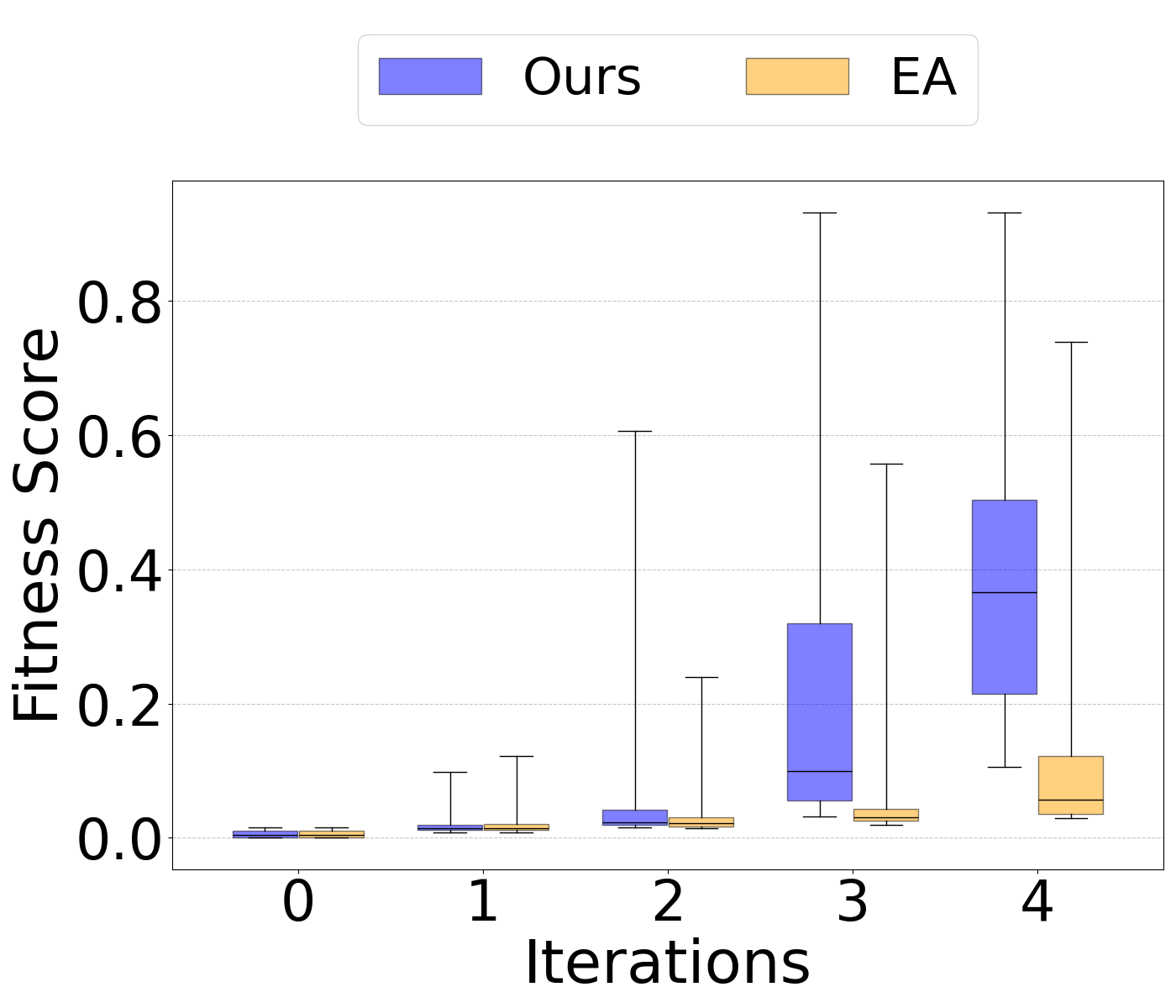} & 
\includegraphics[width=0.3\textwidth]{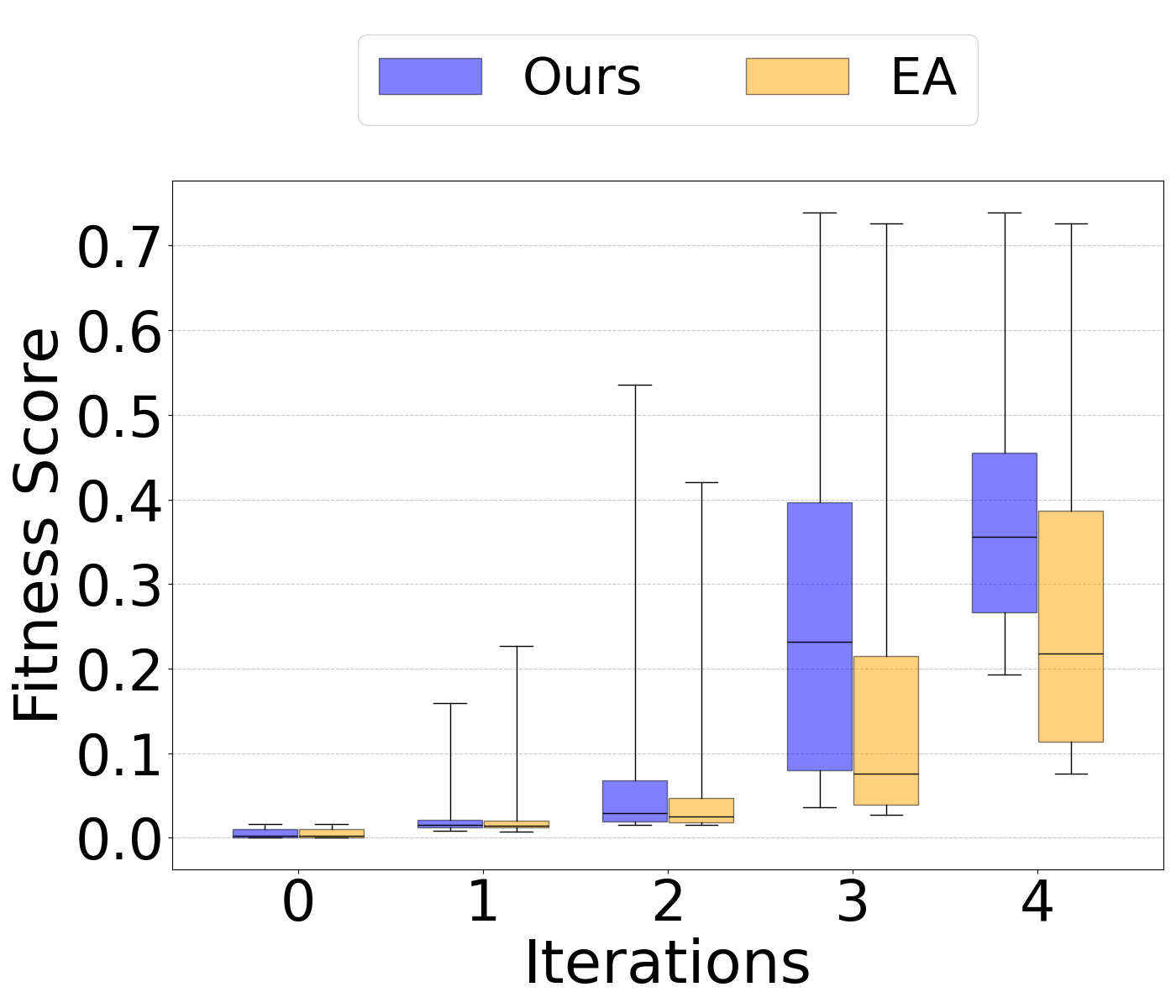} &
\includegraphics[width=0.3\textwidth]{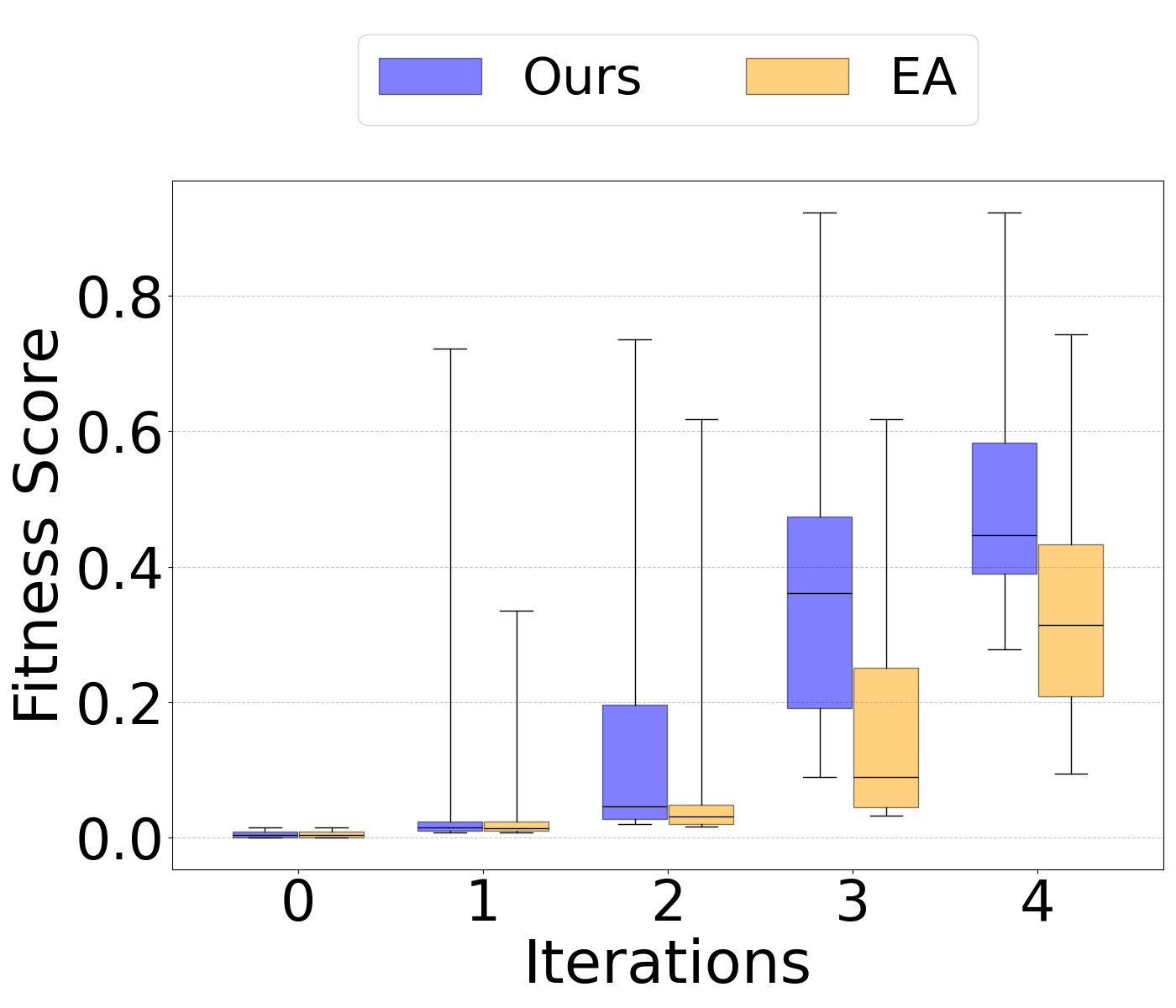} \\
\multicolumn{3}{c}{TrpB}  \\
\includegraphics[width=0.3\textwidth]{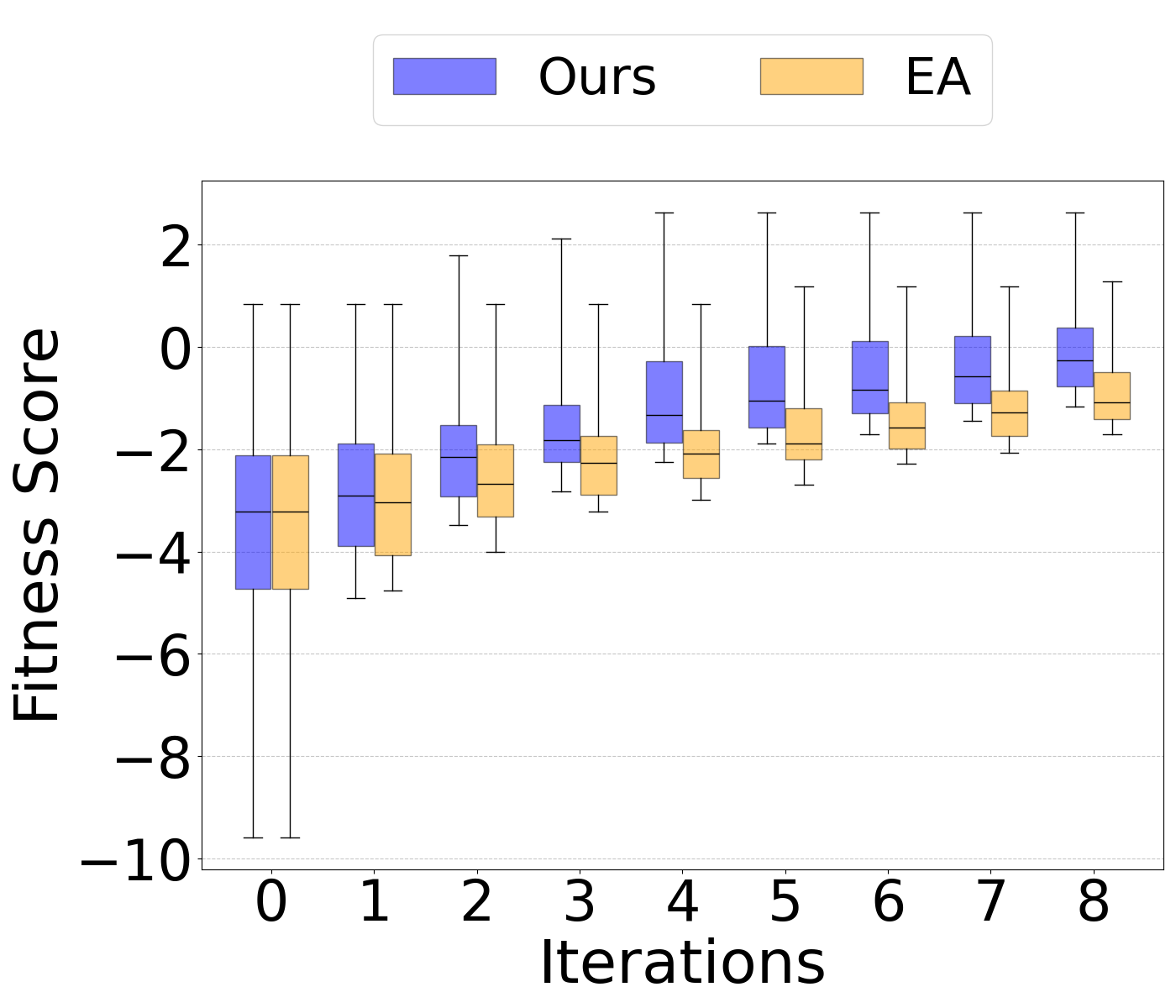} & 
\includegraphics[width=0.3\textwidth]{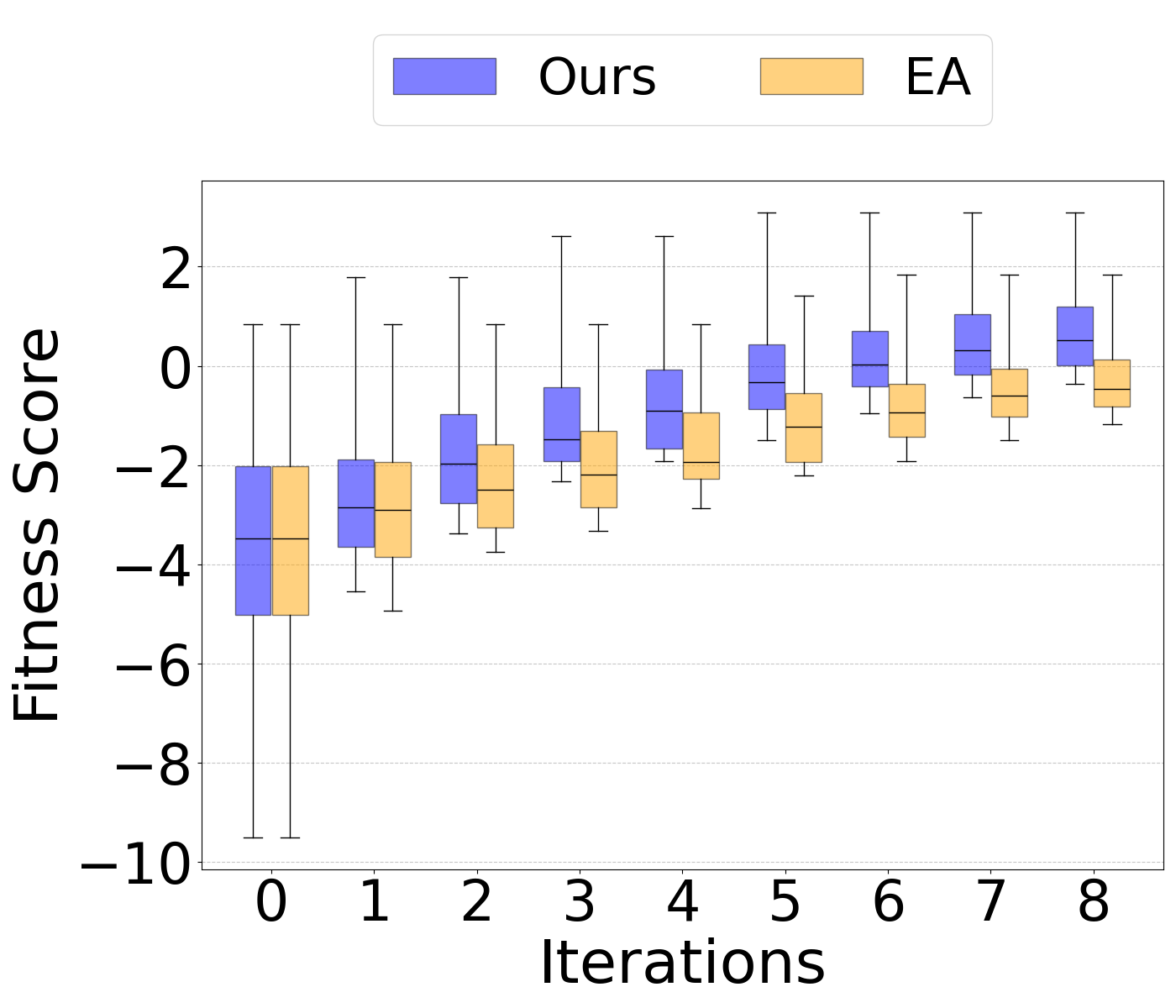} &
\includegraphics[width=0.3\textwidth]{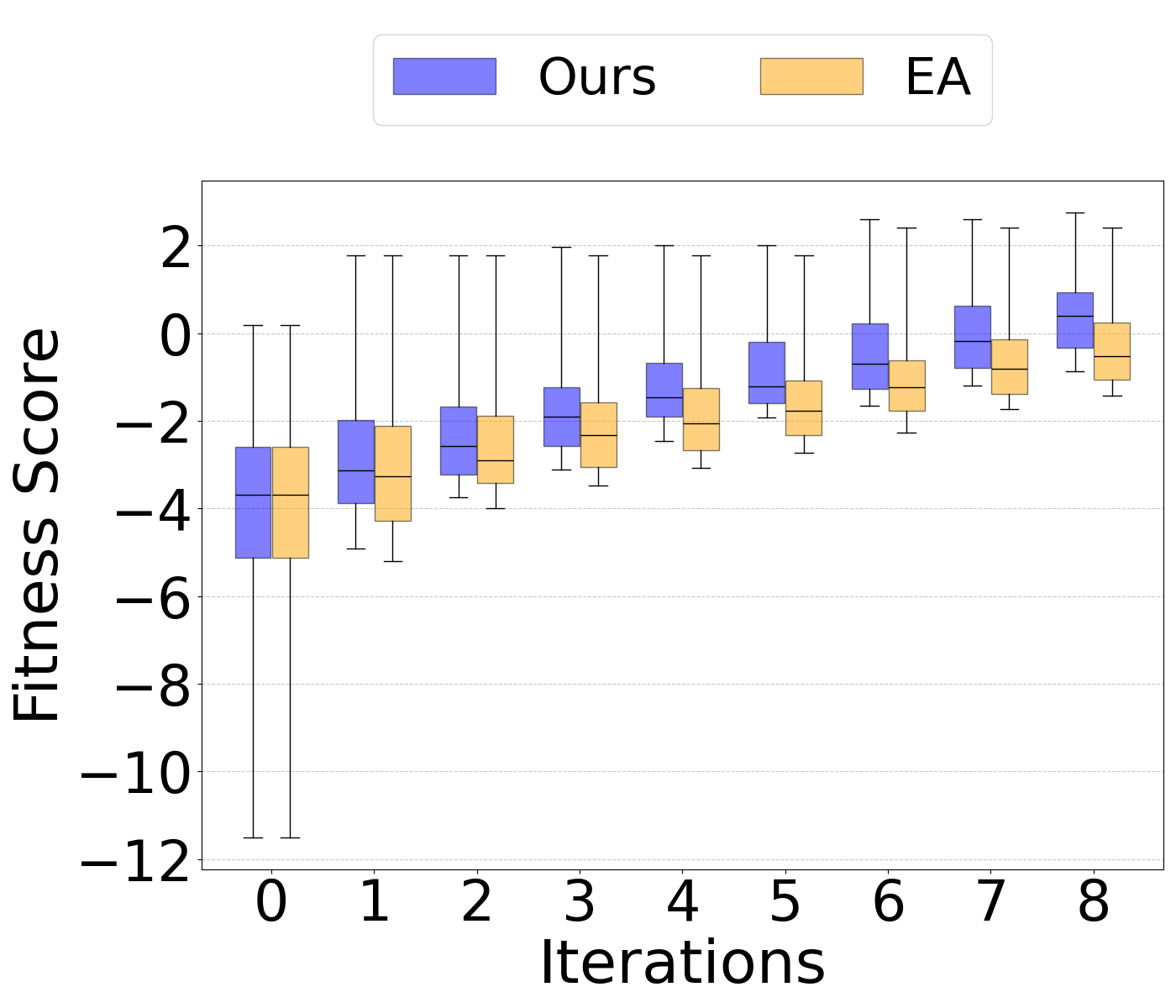} \\
\multicolumn{3}{c}{Syn-3bfo}  \\
\includegraphics[width=0.3\textwidth]{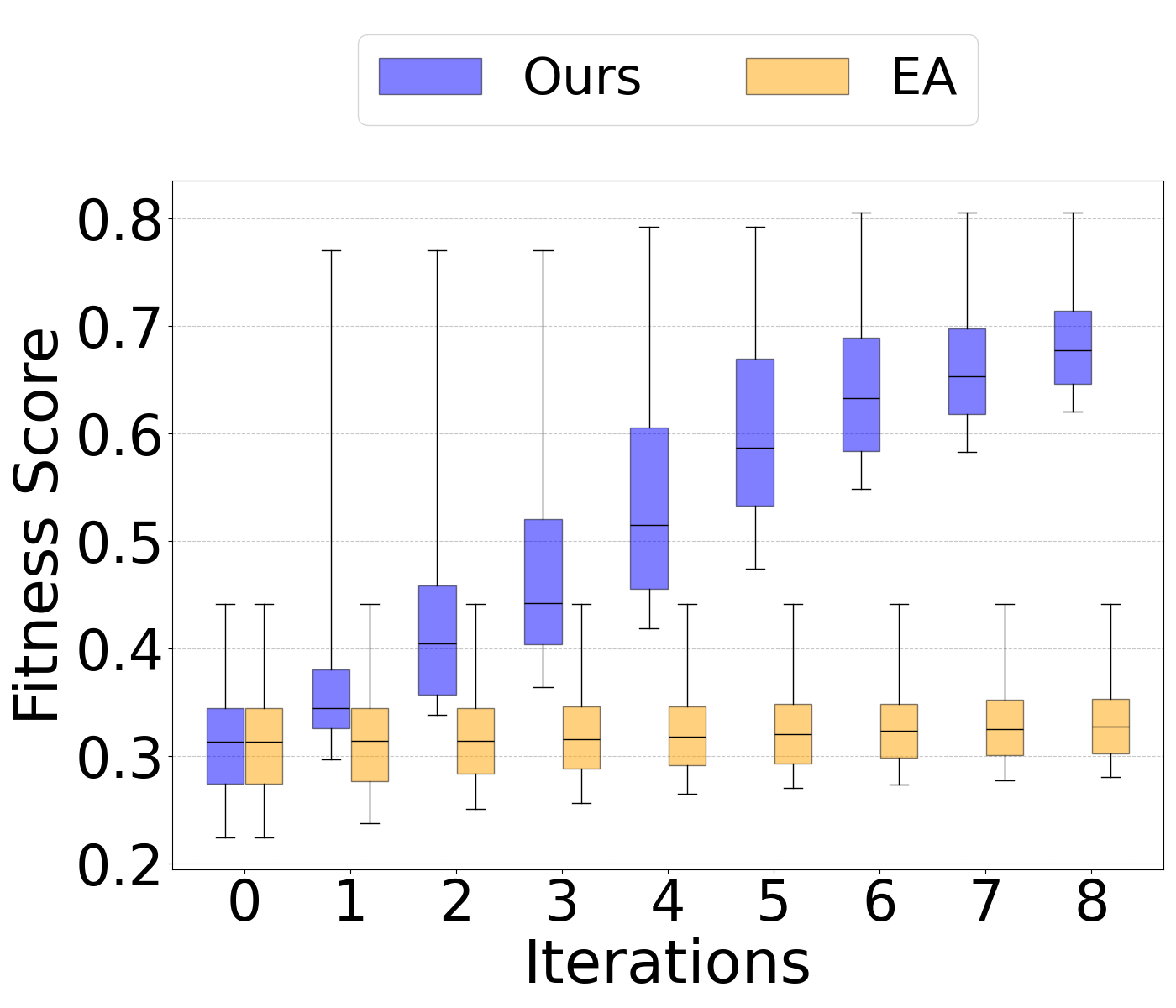} & 
\includegraphics[width=0.3\textwidth]{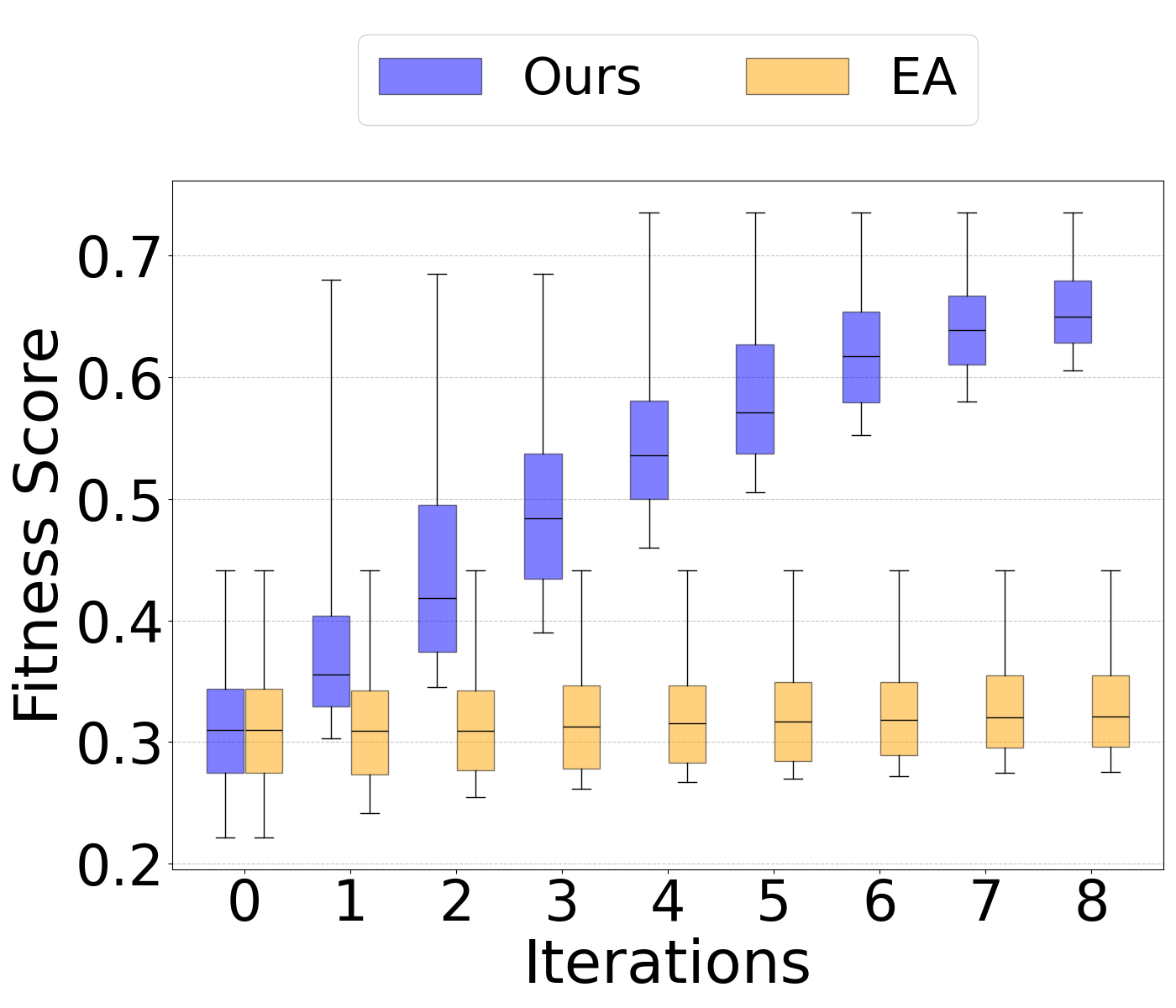} &
\includegraphics[width=0.3\textwidth]{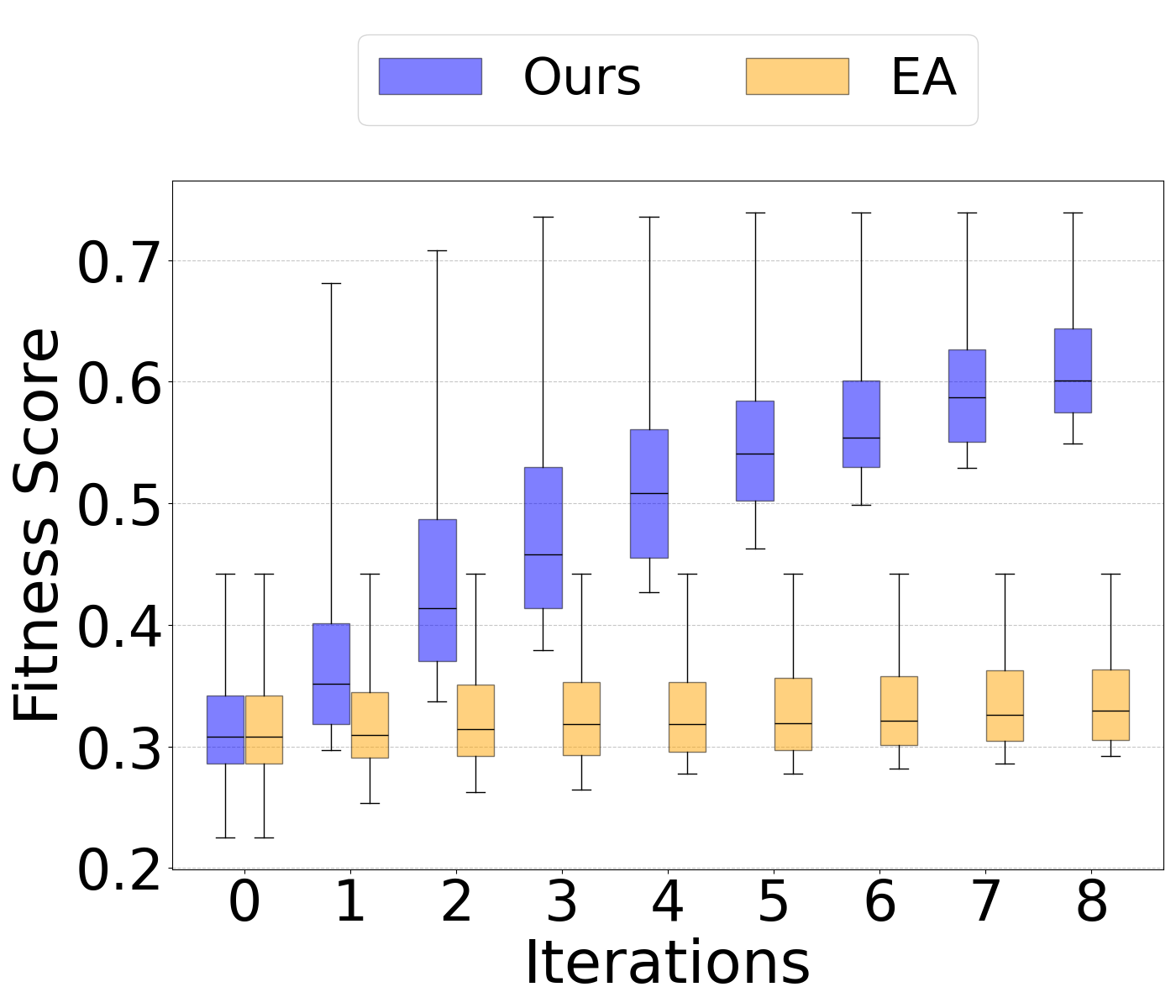} \\
\multicolumn{3}{c}{AAV}  \\
\includegraphics[width=0.3\textwidth]{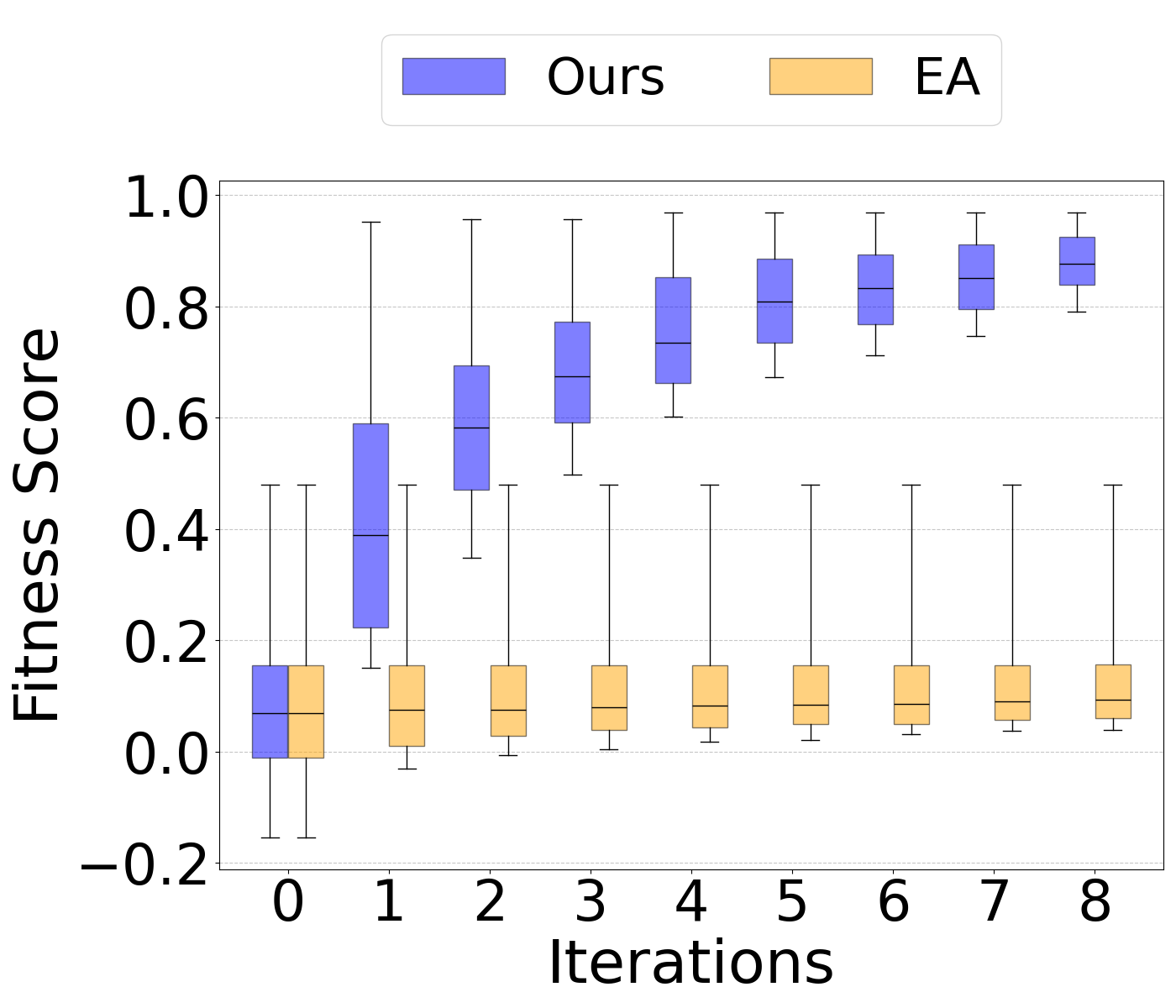} & 
\includegraphics[width=0.3\textwidth]{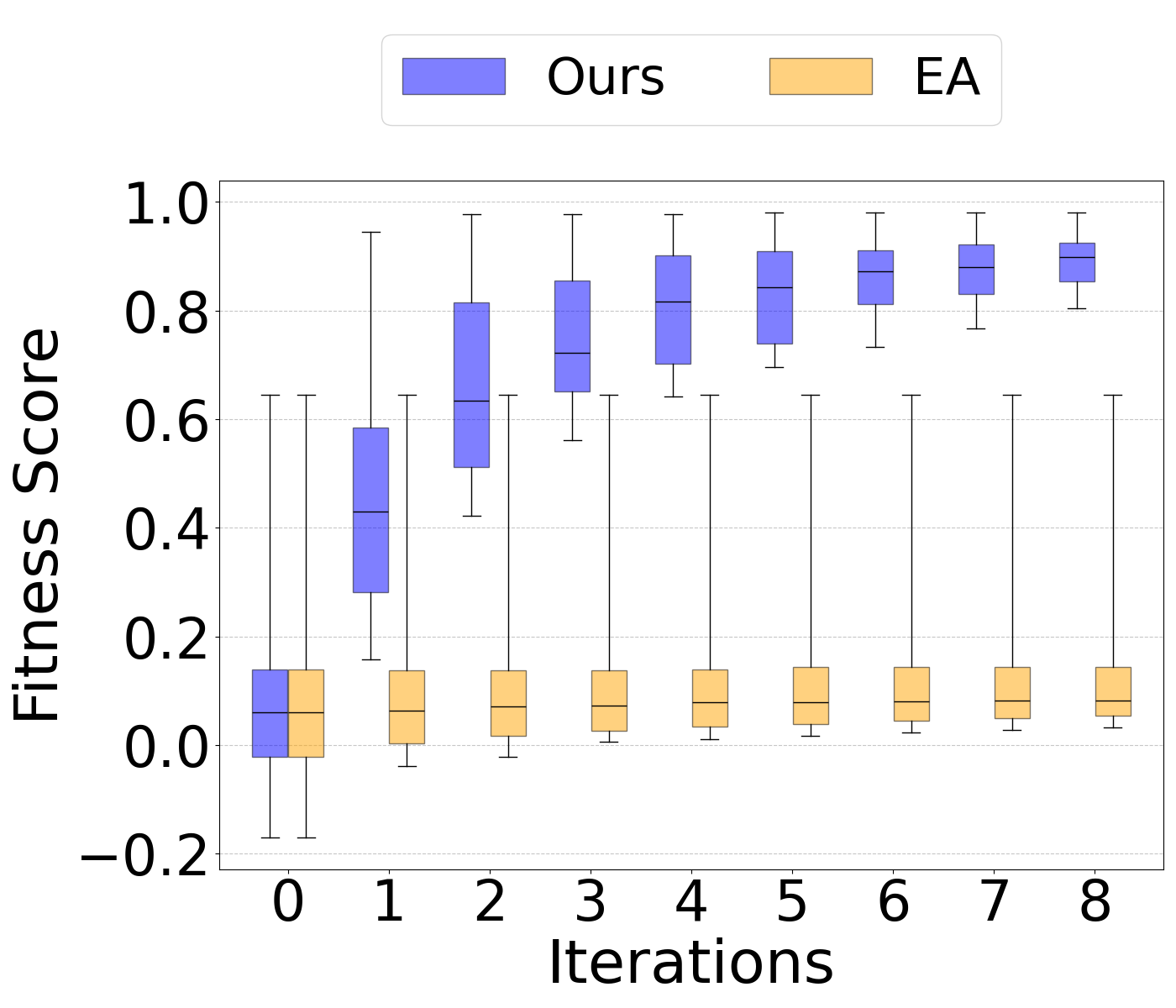} &
\includegraphics[width=0.3\textwidth]{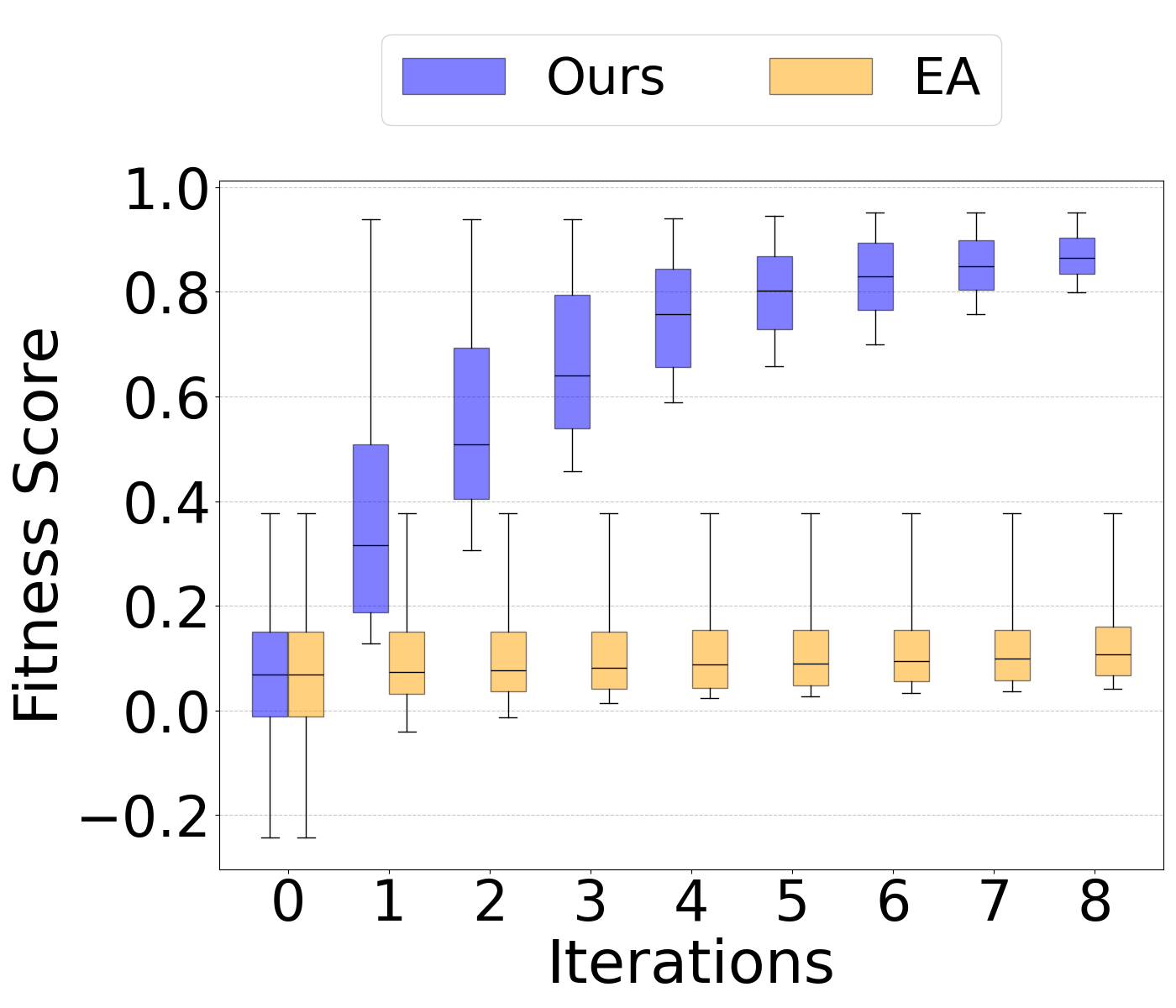} \\
\multicolumn{3}{c}{GFP}  \\
\end{tabular}
\caption{Fitness score across all iterations for five datasets with three random seeds.}
\label{fig:all_single}
\end{figure}

\vspace{-0.5em}
We present additional experiments,show results across three random seeds and datasets in \Cref{fig:all_single}.

\end{document}